\pdfoutput=1

\documentclass[11pt]{article}

\usepackage[]{acl}

\usepackage{times}
\usepackage{latexsym}

\usepackage[T1]{fontenc}

\usepackage[utf8]{inputenc}

\usepackage{microtype}
\usepackage{colortbl}
\usepackage{graphicx}
\usepackage{tablefootnote}
\usepackage{enumitem}
\usepackage[explicit]{titlesec}

\definecolor{darkGray}{gray}{.6}
\definecolor{mediumGray}{gray}{.8}
\definecolor{lightGray}{gray}{.94}
\usepackage{amssymb}
\usepackage{xparse}
\NewDocumentCommand\redheart{}{\includegraphics[scale=0.1]{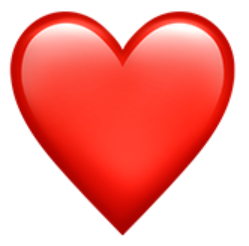}}
\NewDocumentCommand\blueheart{}{\includegraphics[scale=0.1]{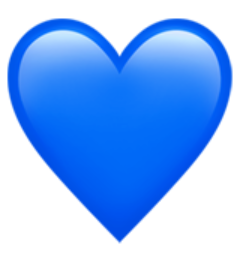}}

\title{Automatic Identification and Classification of Bragging in Social Media}

\author{Mali Jin\textsuperscript{1}, Daniel Preo\c{t}iuc-Pietro\textsuperscript{2}, A. Seza Doğruöz\textsuperscript{3}, Nikolaos Aletras\textsuperscript{1} \\
        \textsuperscript{1}University of Sheffield, \textsuperscript{2}Bloomberg, \textsuperscript{3}Ghent University\\
        \texttt{\{mjin6, n.aletras\}@sheffield.ac.uk} \\
        \texttt{dpreotiucpie@bloomberg.net} \\
        \texttt{as.dogruoz@ugent.be} \\
}

\begin{document}
\maketitle
\begin{abstract}
Bragging is a speech act employed with the goal of constructing a favorable self-image through positive statements about oneself. 
It is widespread in daily communication and especially popular in social media, where users aim to build a positive image of their persona directly or indirectly.
In this paper, we present the first large scale study of bragging in computational linguistics, building on previous research in linguistics and pragmatics.
To facilitate this, we introduce a new publicly available data set of tweets annotated for bragging and their types. We empirically evaluate different transformer-based models injected with linguistic information in (a) binary bragging classification, i.e., if tweets contain bragging statements or not; and (b) multi-class bragging type prediction including not bragging. Our results show that our models can predict bragging with macro F1 up to 72.42 and 35.95 in the binary and multi-class classification tasks respectively.
Finally, we present an extensive linguistic and error analysis of bragging prediction to guide future research on this topic.\footnote{Data is available here: \url{https://archive.org/details/bragging_data}}

\end{abstract}

\section{Introduction}

The desire to be viewed positively is a key driver of human behavior \citep{baumeister1982self,leary1990impression,sedikides1993assessment,tetlock2002social} and creating a positive image often leads to personal rewards \citep{gilmore1989effects,hogan1982socioanalytic,schlenker1980impression}.
Self-presentation strategies are means for individuals to build and establish this positive social image to meet their goals \citep{goffman1978presentation,jones1982toward,jones1990interpersonal,bak2014self}. Bragging (or self-praise) is one of the most common strategies and involves disclosing a positively valued attribute about the speaker or their in-group~\citep{dayter2014self,dayter2018self}.

Social media platforms tend to promote self-presentation tendencies \citep{chen2016anonymity} and allow users to craft an idealized self-image of themselves \citep{chou2012they,michikyan2015can,halpern2017online}. Self-presentation online is predominantly positive \citep{chou2012they,lee2014puts,matley2018not}. Furthermore, self-promotion is acceptable and even desired in certain online contexts \citep{dayter2018self}. This is also amplified by social media platforms through the presence of likes or positive reactions to users' posts \citep{reinecke2014authenticity} which often are used to quantify impact on the platform~\citep{lampos-etal-2014-predicting}.
Bragging in particular was found to be more frequent on social media than face-to-face interactions \citep{ren2020self}.

\renewcommand{\arraystretch}{1.0}
\begin{table*}[!t]
\small
\center
\begin{tabular}{|l|m{0.5\textwidth}|m{0.3\textwidth}|}
\hline
\rowcolor{mediumGray} \bf Type & \bf Definition & \bf Tweet\\
\hline
\rowcolor{lightGray} Achievement & Concrete outcome obtained as a result of the tweet author’s actions. These may include accomplished goals, awards and/or positive change in a situation or status (individually or as part of a group). &\emph{Finally got the offer! Whoop!!}\\
Action & Past, current or upcoming action of the user that does not have a concrete outcome. &\emph{Guess what! I met Matt Damon today!}\\
\rowcolor{lightGray} Feeling & Feeling that is expressed by the user for a particular situation. &\emph{Im so excited that I am back on my consistent schedule. I am so excited for a routine so I can achieve my goals!!}\\
Trait & A personal trait, skill or ability of the user. &\emph{To be honest, I have a better memory than my siblings}\\
\rowcolor{lightGray} Possession & A tangible object belonging to the user. & \emph{Look at our Christmas tree! I kinda just wanna keep it up all year!}\\
Affiliation & Being part of a group  (e.g. family, fanclub, university, team, company etc.) and/or a certain location including living in a city, neighborhood or country. &\emph{My daughter got first place in the final exam, so proud of her!}\\
\rowcolor{lightGray}Not Bragging & The tweet is not about bragging or (a) there is not enough information to determine that the tweet is about bragging; (b) the bragging statements belong to someone other than the author of the tweet; (c) the relationship between the author and people or things mentioned in the tweet are unknown. &\emph{Glad to hear that! Well done Jim!}\\
\hline
\end{tabular}
\caption{Bragging taxonomy together with type definitions and examples of tweets.}
\label{t:braggingExample}
\end{table*}

However, bragging is considered a high risk act \citep{brown1987politeness,holtgraves1990language,van2017praising} and can lead to the opposite effect than intended, such as dislike or decreased perceived competence \citep{jones1982toward,sezer2018humblebragging,matley2018not}. It is, thus, paramount to understand the types of bragging and strategies to mitigate the face-threat introduced by bragging as well as how effective the self-presentation attempt is \citep{herbert1990sex}. Table \ref{t:braggingExample} shows examples of a non-bragging and bragging statements grouped in six types under a taxonomy that we propose in this paper based on previous linguistic research~\cite{dayter2018self,matley2018not}.

Despite its pervasiveness and importance in online communication, bragging has yet to be studied at scale in computational (socio) linguistics.
The ability to identify bragging automatically is important for: (a) linguists to better understand the context and types of bragging through empirical studies~\citep{dayter2014self,ren2020self}; (b) social scientists to analyze the relationship between bragging and personality traits, online behavior and communication strategies \citep{miller1992should,van2017praising,sezer2018humblebragging}; (c) online users to enhance their self-presentation strategies \citep{miller1992should,dayter2018self}; (d) enhancing NLP applications such as intent identification \citep{wen2017jointly} and conversation modeling.


In this paper, we aim to bridge the gap between previous work in pragmatics and the computational study of speech acts. 
Our contributions are:
\begin{itemize}
    \item A new publicly available data set containing a total of 6,696 English tweets annotated with bragging and their types;
    \item Experiments with transformer-based models combined with linguistic features for bragging identification (binary classification) and bragging type classification (seven classes);
    \item A qualitative linguistic analysis of markers of bragging in tweets and the model behavior in predicting bragging.
\end{itemize}

\section{Related Work}

\paragraph{Bragging as a Speech Act} 

Bragging as a speech act is considered a face-threatening act to positive face (i.e. the desire to be liked) under politeness theory \cite{brown1987politeness}. It is directly oriented to the speaker and may threaten their likeability if the bragging is perceived negatively, while also may affect hearer's face by implying that their feelings are not valued by the speaker \citep{matley2018not}. Bragging online plays an important role in self-presentation and its pervasiveness challenges classic politeness theories, such as the modesty maxim \citep{leech2016principles} and the self-denigration maxim \citep{gu1990politeness}. Thus, research in social psychology and linguistics has mostly focused on identifying the pragmatic strategies for bragging that mitigate face threat and their impact of likeability and perceived competence, which the speakers aim to increase with this self-presentation strategy. 

\paragraph{Bragging Strategies} 

Modest and sincere self-presentation styles are more likely to be perceived positively \citep{sedikides2007}. Bragging framed as mere information-sharing, but with positive connotation to the speaker, can make the speaker be perceived as more likeable \citep{miller1992should}. It can also be perceived negatively and causes greater aggression when it involves boasting, elements of competitiveness, use of superlatives and explicit comparisons to others \citep{miller1992should,hoorens2012hubris,scopelliti2015you,matley2018not}. In addition, competence related statements are more likely to be negatively perceived than those based on warmth (e.g. the ability to form connections with others) \citep{van2017praising}. Common mitigation strategies include speaker's attempts to deny compliments, shifting focus to persons closely related to them, reframing bragging as praise from a third party, admitting the bragging act through disclaimers (e.g. using \#brag) or expressing it as a complaint \cite{wittels2011humblebrag,sezer2018humblebragging}, question, narration or sharing \cite{dayter2018self,matley2018not,ren2020self}. The success of self-presentation strategies are also impacted by the social context \citep{tice1995modesty} or speaker identity \citep{paramita2021benefits}.

\paragraph{Analysis of Bragging}

Bragging has been studied in the context of a small ballet community \citep{dayter2014self}, a pick-up artist forum \citep{rudiger2020manbragging} and a small set of WhatsApp conversations \citep{dayter2018self}. On social media, \citet{matley2018not} studied the functional use of hashtags (e.g. \#brag, \#humblebrag) in Instagram posts, \citet{tobback2019telling} examined bragging strategies on LinkedIn, \citet{ren2020self} investigated bragging and its pragmatic functions in Chinese social media and \citet{matley2020isn} studied impact of mitigating bragging through irony showing that bragging was negatively perceived. However, all these studies rely on manual analyses of small data sets (e.g. $<$300 posts). 

\paragraph{Speech Acts in NLP} 

Speech acts have been studied in NLP with examples 
including politeness \citep{danescu-niculescu-mizil-etal-2013-computational}, complaints \citep{preotiuc-pietro-etal-2019-automatically, jin-aletras-2020-complaint, jin-aletras-2021-modeling}, humor \citep{yang-etal-2021-choral}, parody \citep{maronikolakis-etal-2020-analyzing}, irony \citep{bamman2015contextualized}, deception \citep{chen-etal-2020-acoustic} and self-disclosure \citep{bak-etal-2012-self,levontin2017negative,ravichander-black-2018-empirical}. Self-disclosure is closer to bragging as it is related to revealing personal information about oneself. It is usually employed to improve or maintain relationships \citep{bak-etal-2012-self} as measured through conversation frequency \citep{bak2014sdtm}. On the other hand, bragging is about aspects that are positively valued by the audience with the goal of improving the speaker's self-image. \citet{bak2014self} aim to predict different levels of self-disclosure statements, from general to sensitive; while \citet{wang-etal-2021-bragging} examine gender differences in self-promotion by Congress members on Twitter. Bragging also involves in some cases possessions \citep{chinnappa-blanco-2018-mining}.



\section{Bragging Data}


\subsection{Bragging Definition \& Types}
\label{sec:category}


\paragraph{Definition} Bragging is a speech act which explicitly or implicitly attributes credit to the speaker for some \textit{good} (e.g.possession, skill) that is positively valued by the speaker and their audience \citep{dayter2014self}. A bragging statement should clearly express what the author is bragging about.

\paragraph{Types} 
We generalize and extend the bragging types based on the definitions by \citet{dayter2018self} and \citet{matley2018not}. The former summarizes them as accomplishments and some aspects of self; while the latter includes everyday achievements (e.g. cooking) and personal qualities. We divide the `some aspects of self' category into two categories, namely `Possession' and `Trait' respectively. We also add an `Affiliation' category for bragging involving a group to which the speaker belongs. In total, we consider six bragging types and a non-bragging category. Table \ref{t:braggingExample} shows the definitions of each type.

\paragraph{Classification Tasks} 
Given the taxonomy above, we define two classification tasks: (i) \emph{binary} bragging prediction (i.e. if a tweet contains a bragging statement or not); and (ii) seven-way \emph{multiclass} classification for predicting if a tweet contains one of the six bragging types or no bragging at all.



\subsection{Data Collection}

To the best of our knowledge, there is no other data set available for our study. We use Twitter for data collection as tweets are openly available for research and widely used in other related tasks, e.g. predicting sentiment~\citep{rosenthal2017semeval}, affect~\citep{SemEval2018Task1}, sarcasm~\citep{bamman2015contextualized}, stance~\citep{mohammad2016semeval}. 

\paragraph{Random Sampling}~We select tweets for annotation by randomly sampling from the 1\% Twitter feed one day per month from January 2019 to December 2020 (approximately 10k tweets per day) to ensure diversity using the Premium Twitter Search API for academic research.\footnote{\url{https://tinyurl.com/2p8wnure}}  

\paragraph{Keyword-based Sampling}~To give a model access to more positive examples of bragging statements for training, we use a keyword-based sampling method that increases the hit rate of bragging, following previous work on labeling infrequent linguistic phenomena, e.g. irony~\citep{SemEval2018Task1} or hate speech~\cite{waseem-hovy-2016-hateful}. 

We build queries based on indicators of positive self-disclosure (e.g. \emph{I, just}) \cite{dayter2018self} and stylistic indicators, e.g. positive emotion words, present tense verbs \cite{bazarova2013managing}. As the frequency of these keywords is high, we construct multi-word queries consisting of a personal pronoun and an indicator. In addition, we use a short list of curated bragging-related hashtags.\footnote{The queries are: \{[\emph{I, proud}], [\emph{I, glad}], [\emph{I, happy}], [\emph{I, best}], [\emph{I, amazed}], [\emph{I, amazing}], [\emph{I, excellent}], [\emph{I, just}], [\emph{I'm, proud]}, [\emph{I'm, glad}], [\emph{I'm, happy}], [\emph{I'm, best}], [\emph{I'm, amazed}], [\emph{I'm, amazing}], [\emph{I'm, excellent}], [\emph{me, proud}], [\emph{my, best}], \#\emph{brag}, \#\emph{bragging}, \#\emph{humblebrag}, \#\emph{humble}, \#\emph{braggingrights}\}.} After annotating 1,000 tweets, we compute the percentage of bragging tweets for each keyword and remove from sampling tweets with less than 5\% (i.e. [\emph{I, amazed}], [\emph{I'm, amazing}], [\emph{I'm, best}], [\emph{my, best}], [\emph{I, excellent}], \#\emph{humble}).

We initially collected around 6K and 368K tweets using hashtags and multi-word queries respectively. We obtain over 9k tweets by keeping all tweets collected using hashtags and sample 1\% from those collected using multi-word queries to balance the two types.

\paragraph{Data Filtering} After collecting tweets, we exclude those with duplicate or no meaningful textual content (e.g. only @-mentions or images). We only focus on English posts and filter out non-English ones using the language code provided by Twitter. We also exclude retweets and quoted tweets, as these do not typically express the thoughts of the user who retweeted them. Moreover, we exclude 131 tweets containing a URL in the text because these were related to advertisements based on initial results from our annotation calibration rounds. This resulted in a total of 6,696 tweets which is of similar size with data sets recently released for social NLP~\cite{oprea-magdy-2020-isarcasm, chung-etal-2019-conan, beck-etal-2021-investigating, mendelsohn-etal-2021-modeling}.

\renewcommand{\arraystretch}{1.0}
\begin{table}[!t]
\center
\resizebox{0.48\textwidth}{!}{
\begin{tabular}{|l|c|c|c|}
\hline
\rowcolor{darkGray} \bf Label & \bf Training set & \bf Dev/Test set &\bf All \\ \hline
\rowcolor{darkGray} \bf  & \bf (Keyword sampling) & \bf (Random sampling) &\bf  \\ \hline
\hline
\rowcolor{lightGray} Binary &&& \\
Bragging & 544 (16.09\%) & 237 (7.15\%) & 781 (11.66\%)  \\
Not Bragging & 2838 (83.91\%) & 3077 (92.85\%) & 5915 (88.34\%) \\
\hline
\rowcolor{lightGray} Multi-class &&& \\
Achievement & 166 (4.91\%) & 71 (2.14\%) & 237 (3.54\%) \\
Action & 127 (3.76\%) & 58 (1.72\%) & 185 (2.76\%) \\
Feeling & 39 (1.15\%) & 27 (0.82\%) & 66 (0.99\%) \\
Trait & 91 (2.69\%) & 48 (1.45\%) & 139 (2.08\%) \\
Possession & 58 (1.72\%) & 28 (0.84\%) & 86 (1.28\%) \\
Affiliation & 63 (1.86\%) & 5 (0.15\%) & 68 (1.01\%)\\
Not Bragging & 2838 (83.91\%) & 3077 (92.85\%) & 5915 (88.34\%) \\
\hline
\rowcolor{lightGray} Total & 3382 & 3314 & 6696 \\
\hline
\end{tabular}}
\caption{Bragging data set statistics.}
\label{t:statistics}
\end{table}

\subsection{Annotation and Quality Control Process}

We manually annotate tweets for providing a solid benchmark and foster future research. 
All authors of the paper have significant experience in linguistic annotation. We run three calibration rounds of 100 tweets each, where all annotated all tweets and discussed disagreements, until a Krippendorf's Alpha above 0.80 in the seven-class task was reached.

To monitor quality, a subset of 1,564 tweets were annotated by two annotators or more in case of disagreements. If a tweet fits into multiple bragging types, we assign the more prominent one.\footnote{For example, we annotate \textit{``New car\checkmark New crib\checkmark New barbershop\checkmark 20 years young''} as `Possession' because bragging is mostly about possessions (\textit{crib}, \textit{car}, \textit{barbershop}).} The annotation is based only on the actual text of the tweet without considering additional modalities (e.g. images), context or replies. This is similar to the information available to predictive models during training. We selected the final label as the majority vote and a final label was assigned after consensus in cases of three different votes.\footnote{We experimented on training models using the subset annotated by a single annotator compared to multiple annotators and find no significant differences (see Appendix~\ref{s:annotation_results}).} The full task guidelines, examples and interface are presented in Appendix~\ref{s:interface}.

The inter-annotator agreement between two annotations of all tweets is: (a) percentage agreement: 89.03; (b) Krippendorf's Alpha \citep{krippendorff2011computing} (7-class): 0.840; (c) Krippendorf's Alpha (binary): 0.786. Agreement values are between the upper part of the \textit{substantial} agreement band and the \textit{perfect} agreement band~\citep{artstein2008inter}. 
The final data set consists of 6,696 tweets with one of the seven classes. Before annotation, the keyword-based and randomly sampled tweets were shuffled to not induce frequency bias. Data set statistics are shown in Table \ref{t:statistics}, including statistics across the two sampling strategies. The model performance curve by varying the training set size indicates that annotating more data is not likely to lead in substantial improvements in bragging prediction (see Figure \ref{fig:learning_curve} in Appendix).



\renewcommand{\arraystretch}{1.1}
\begin{table}[!t]
\small
\center
\resizebox{0.48\textwidth}{!}{
\begin{tabular}{|l|c|c|}
\hline
\rowcolor{darkGray} \bf Class & \bf Self-disclosure (\%) & \bf Non-self-disclosure (\%) \\
\hline
Bragging & 31.63 & 68.37 \\
\rowcolor{lightGray} Non-bragging & 24.04 & 75.96 \\
\hline
Achievement & 31.65 & 68.35 \\
\rowcolor{lightGray} Action & 27.57 & 72.43 \\
Feeling & 31.82 & 68.18\\
\rowcolor{lightGray} Trait & 36.69 & 63.31 \\
Possession & 29.07 & 70.93\\
\rowcolor{lightGray} Affiliation & 35.29 & 64.71 \\
Non-bragging & 24.04 & 75.96  \\
\hline
\rowcolor{lightGray} Total & 24.93 & 75.07 \\
\hline
\end{tabular}}
\caption{Percentages of self-disclosure class across bragging classes}
\label{t:self-disclosure}
\end{table}

\subsection{Self-disclosure in Bragging}
We conduct an analysis of the relationship between self-disclosure and bragging as they are closely related. We use self-disclosure lexicon by \citet{bak2014self} to assign each tweet in our data set a label (i.e. self-disclosure or non-self-disclosure). The percentages of self-disclosure across each bragging type are shown in Table \ref{t:self-disclosure}. We also used self-disclosure models as a predictor for bragging in early experimentation but the results are omitted due to the low performance.

\subsection{Data Splits}
We use the keyword sampled data for training and the random data for development and testing (in the ratio of 2:8) because the latter is representative of the real distribution of tweets (see Table \ref{t:statistics}). 

\section{Predictive Models}

We evaluate vanilla transformer-based models~\cite{vaswani2017attention} and further leverage external linguistic information to improve them. 

\paragraph{BERT, RoBERTa and BERTweet}
We experiment with Bidirectional Encoder Representations from Transformers (BERT; \citet{devlin-etal-2019-bert}), RoBERTa \citep{liu2019roberta} and BERTweet \citep{nguyen-etal-2020-bertweet}. RoBERTa is a more robust variant of BERT that obtains better results on a wide range of tasks. BERTweet is pre-trained on English tweets using RoBERTa as basis and achieves better performance on Twitter tasks \citep{nguyen-etal-2020-bertweet}. We fine-tune BERT, RoBERTa and BERTweet for binary and multiclass bragging prediction by adding a classification layer that takes the [CLS] token as input.

\paragraph{BERTweet with Linguistic Features}

We inject linguistic knowledge that could be related to bragging to the BERTweet model with a similar method proposed by \citet{jin-aletras-2021-modeling},\footnote{Early experimentation with simply concatenating or applying attention resulted in lower performance.} that was found to be effective on complaint severity classification, a related pragmatics task. The method is adapted from \citet{rahman2020integrating}, which integrates multimodal information (e.g. audio, visual) in transformers using a fusion mechanism called Multimodal Adaption Gate (MAG). MAG integrates multimodal information to text representations in transformer layers using an attention gating mechanism for modality influence controlling. We first expand vectors of linguistic information to a comparable size to the embeddings fed to the pre-trained transformer. We, then, use MAG to concatenate contextual and linguistic representations after the embedding layer of the transformer similar to \citet{rahman2020integrating}. 
The output is sent to a pre-trained BERTweet encoder for fine-tuning followed by an output layer. 

We experiment with these linguistic features:  


%
%
\begin{itemize}
    \item  {\bf NRC:}~The NRC word-emotion lexicon contains a list of English words mapped to ten categories related to emotions and sentiment \cite{mohammad2013crowdsourcing}. We represent each tweet as a 10-dimensional vector where each element is the proportion of tokens belonging to each category.

    \item {\bf LIWC:}~Linguistic Inquiry and Word Count \cite{pennebaker2001linguistic} is a dictionary-based approach to count words in linguistic, psychological and topical categories. We use LIWC 2015 to represent each tweet as a 93-dimensional vector.
    \item{\bf Clusters:}~We use Word2Vec clusters proposed by \citet{preoctiuc2015studying} to represent each tweet as a 200-dimensional vector over thematic subjects.
\end{itemize}

\section{Experimental Setup}

\paragraph{Text Processing}

We pre-process text by lowercasing, replacing all username mentions with placeholder tokens \emph{@USER} and emojis with words using demojize.\footnote{\url{https://pypi.org/project/emoji/}} We also remove hashtags that are used as keywords (e.g. \emph{\#brag}) in data collection. Finally, we tokenize the text using TweetTokenizer.\footnote{\url{https://www.nltk.org/api/nltk.tokenize.html}} 

\renewcommand{\arraystretch}{1.2}
\setlength{\tabcolsep}{3pt}
\begin{table*}[!t]
\scriptsize
\center
\begin{tabular}{|l|c|c|c||c|c|c|}
\hline
\rowcolor{darkGray} \bf Model & \bf Precision & \bf Recall & \bf Macro-F1 & \bf Precision & \bf Recall & \bf Macro-F1\\
\rowcolor{mediumGray} & \multicolumn{3}{|c||}{Bragging Classification (Binary)} & \multicolumn{3}{|c|}{Bragging and Type Classification (7 class)} \\
\hline
Majority Class & 46.42 & 50.00 & 48.15 & 13.26 & 14.29 & 13.76 \\
\rowcolor{lightGray} LR-BOW & 54.53 & 63.16 & 52.68 & 18.52 & 20.02 & 18.59 \\
BiGRU-Att & 55.93 $\pm$ 1.53 & 51.41 $\pm$ 0.47 & 51.29 $\pm$ 1.40 & 18.32 $\pm$ 0.10 & 26.16 $\pm$ 3.41 & 19.19 $\pm$ 0.31\\
\hline
\rowcolor{lightGray} BERT & 64.24 $\pm$ 1.40 & 65.91 $\pm$ 3.32 & 64.58 $\pm$ 0.80 & 24.16 $\pm$ 1.15  & 39.66 $\pm$ 4.84  & 26.85 $\pm$ 0.81 \\
RoBERTa & 66.53 $\pm$ 0.29 & 68.43 $\pm$ 2.05 & 67.34 $\pm$ 1.02 & 28.99 $\pm$ 0.61 & 45.90 $\pm$ 3.59 & 32.82 $\pm$ 0.65\\
\rowcolor{lightGray} BERTweet & 70.43 $\pm$ 0.16 & \textbf{72.62} $\pm$ 0.89 & 71.44 $\pm$ 0.43 & 30.82 $\pm$ 0.75 & 47.25 $\pm$ 2.68 & 34.86 $\pm$ 0.79 \\
\hline
BERTweet-NRC & 72.89 $\pm$ 1.26 & 70.95  $\pm$ 0.96 & 71.80  $\pm$ 0.49 & 30.95 $\pm$ 0.54 & \textbf{47.98} $\pm$ 1.12  & 34.36 $\pm$ 0.19 \\
\rowcolor{lightGray} BERTweet-LIWC & 72.65 $\pm$ 0.20 & 72.21 $\pm$ 0.43 & $\textbf{72.42}^\dagger$ $\pm$ 0.31 & 32.06 $\pm$ 2.42 & 46.68 $\pm$ 7.45 & 34.83 $\pm$ 0.79 \\
BERTweet-Clusters & 71.26 $\pm$ 2.27 & 72.53 $\pm$ 1.91 & 71.60 $\pm$ 0.21 & \textbf{32.51} $\pm$ 1.36 & 46.97 $\pm$ 2.36 & \textbf{35.95} $\pm$ 0.54 \\
\hline
\end{tabular}
\caption{Macro precision, recall and F1-Score (± std. dev. for 3 runs) for bragging prediction (binary and multiclass). Best results are in bold. $\dagger$ indicates significant improvement over BERTweet (t-test, p$<$0.05).}
\label{t:multiresults}
\end{table*}

\paragraph{Baselines}

\paragraph{Majority Class:} As a first baseline, we label all tweets with the label of the majority class.

\paragraph{LR-BOW:} We train a Logistic Regression with bag-of-words using L2 regularization.

\paragraph{BiGRU-Att:} We also train a bidirectional Gated Recurrent Unit (GRU) network \citep{cho-etal-2014-learning} with self-attention \citep{tian2018attention}. Tokens are first mapped to GloVe embeddings \citep{pennington2014glove} and then passed to a bidirectional GRU. Subsequently, its output is passed to a self-attention layer and an output layer for classification.

\paragraph{Hyperparameters}
For {\bf BiGRU-Att}, we use 200-dimensional GloVe embeddings \citep{pennington2014glove} pre-trained on Twitter data. The hidden size is $h$ = 128 where $h \in$ \{64, 128, 256, 512\} with dropout $d$ = .2, $d \in$ \{.2, .5\}. We use Adam optimizer \citep{kingma2015method} with learning rate $l$ = 1e-2, $l \in$ \{1e-3, 1e-2, 1e-1\}. For {\bf BERT}, {\bf RoBERTa} and {\bf BERTweet}, we use the base cased model (12 layers and 109M parameters, 12 layers and 125M parameters and 12 layers and 135M parameters accordingly) and fine-tune them with learning rate $l$ = 3e-6, $l \in$ \{1e-4, 1e-5, 5e-6, 3e-6, 1e-6\}. For {\bf BERTweet with linguistic features}, we project these to vectors of size $l_{NRC}$ = 200, $l_{LIWC}$ = 400, $l_{Clusters}$ = 768, $l \in$ \{10, 93, 200, 400, 600, 768\}. For MAG, we use the default parameters from \citet{rahman2020integrating}. 
For \textit{multi-class classification}, we apply class weighting due to the imbalanced data and set the training epoch to $n$ = 40, $n \in$ \{15, 20, 25, 30, 35, 40, 45, 50, 55, 60,\}. The maximum sequence length is set to 50 covering 95\% of tweets in the training set. We use a batch size of 32.

\paragraph{Training and Evaluation}
We train each model three times using different random seeds and report the mean Precision, Recall and F1 (macro). We apply early stopping during training based on the dev loss. 
The experiments with linguistic features are performed with the best pre-trained transformer in each of the two classification tasks.

\section{Results}

\paragraph{Binary Bragging Classification}

Table \ref{t:multiresults} (left) shows the predictive performance of all models on predicting bragging (i.e. binary classification). Overall, BERTweet models with linguistic information achieve better overall performance. Transformer models perform substantially above the \emph{majority class baseline} (+23.29 F1) and above {\it Logistic Regression} (+18.76). {\it  BERTweet} (71.44 F1) performs better than {\it  BERT} (64.58 F1) and {\it RoBERTa} (67.34 F1), which illustrates the advantage of pre-training on English tweets for this task. 

Performance is further improved (+0.98 F1) by using {\it LIWC features} alongside BERTweet, which indicates that injecting extra linguistic information benefits bragging identification. We speculate that this is because a bragging statement usually contains particular terms (e.g. personal pronouns, positive terms) or involves at least one certain aspect or theme (e.g. reward or property), which can be captured by linguistic features (e.g. feature \textit{I} and \textit{ACHIEVE} in LIWC). Combining lexicons lead to worse results than using a single one, so we refrain from reporting these results for clarity. 

\paragraph{Multi-class Bragging Classification}

Table~\ref{t:multiresults} (right) shows the predictive performance of all models on multiclass bragging type prediction including not bragging. We again find that pre-trained transformers substantially outperform the \textit{majority class baseline} (+21.1 F1) and \textit{logistic regression} (+16.27 F1). In line with the binary results, we find that \textit{BERTweet} (34.86 F1) performs best out of all transformers. \textit{BERTweet-Clusters} outperforms all models (35.95 F1), which indicates that topical information helps to identify different types of bragging. Each bragging type might be particularly specialized to certain topics (e.g. \textit{weight loss} in `Achievement' category). 

\setlength{\tabcolsep}{3pt}
\begin{table*}[!t]
\center
\small
\resizebox{1\textwidth}{!}{
\renewcommand{\arraystretch}{1.1}
\begin{tabular}{|l|c|l|c||l|c|l|c|l|c|l|c|l|c|l|c|}
\hline
\rowcolor{darkGray} \multicolumn{2}{|c|}{\bf Bragging} & \multicolumn{2}{|c||}{\bf Non-Bragging} & \multicolumn{12}{|c|}{\bf Bragging type}  \\
\hline
\rowcolor{darkGray} \multicolumn{4}{|c||}{}  & \multicolumn{2}{|c|}{\bf Achievement} & \multicolumn{2}{|c|}{\bf Action} & \multicolumn{2}{|c|}{\bf Feeling} & \multicolumn{2}{|c|}{\bf Trait} & \multicolumn{2}{|c|}{\bf Possession} & \multicolumn{2}{|c|}{\bf Affiliation} \\
\hline
\rowcolor{mediumGray} Feature & r & Feature & r & Feature & r & Feature & r & Feature & r & Feature & r & Feature & r & Feature & r\\
\hline
\rowcolor{darkGray} \multicolumn{16}{|l|}{\bf Unigrams and LIWC} \\
\hline
AUTHENTIC & 0.149 & CLOUT & 0.109 & FOCUSPAST & 0.200 & get & 0.146 & happy & 0.228 & APOSTRO & 0.197 & own & 0.211 & FAMILY & 0.276 \\
\rowcolor{lightGray} my & 0.127 & YOU & 0.089 & Number & 0.157 & trip & 0.128 & POSEMOE & 0.218 & COGPROC & 0.181 & buy & 0.175 & CLOUT & 0.271 \\
I & 0.122 & DISCREP & 0.078 & Analytic & 0.153 & RELATIV & 0.119 & \redheart{}{} & 0.191 & FOCUSPRESENT & 0.179 & bought & 0.149 & proud & 0.263 \\
\rowcolor{lightGray} TONE & 0.104 & NEGEMO & 0.077 & finished & 0.150 & ready & 0.114 & blessed & 0.190 & cute & 0.159 & car & 0.146 & rights & 0.215 \\
FOCUSPAST & 0.102 & SOCIAL & 0.076 & 3 & 0.133 & him & 0.114 & AFFECT & 0.184 & PRONOUN & 0.157 & bedroom & 0.144 & SOCIAL & 0.209 \\
\rowcolor{lightGray} WC & 0.100 & FOCUSPRESENT & 0.070 & WORK & 0.132 & happen & 0.105 & feels & 0.176 & take & 0.143 & extra & 0.144 & amazing & 0.205 \\
RELATIV & 0.090 & INFORMAL & 0.056 & managed & 0.130 & FOCUSFUTURE & 0.105 & love & 0.169 & COMPARE & 0.143 & xr & 0.142 & \blueheart{}{} & 0.197 \\
\rowcolor{lightGray} TIME & 0.081 & COGPROC & 0.056 & over & 0.129 & fun & 0.102 & sunrise & 0.166 & ANGER & 0.138 & macbook & 0.055 & law & 0.185 \\
during & 0.078 & ANGER & 0.056 & under & 0.119 & gave & 0.097 & weighted & 0.162 & I & 0.137 & new & 0.139 & team & 0.182 \\
\rowcolor{lightGray} ACHIEVE & 0.075 & just & 0.054 & beat & 0.112 & hours & 0.096 & july & 0.159 & if & 0.137 & afford & 0.139 & OTHERP & 0.181\\
PREP & 0.073 & your & 0.052 & race & 0.104 & before & 0.095 & time & 0.159 & SWEAR & 0.134 & PERIOD & 0.106 & words & 0.164 \\
\rowcolor{lightGray} managed & 0.072 & IPRON & 0.051 & office & 0.103 & sitting & 0.095 & truly & 0.156 & am & 0.133 & HOME & 0.105 & teams & 0.164 \\
REWARD & 0.069 & ? & 0.043 & possible & 0.103 & VERB & 0.094 & BIO & 0.147 & PPRON & 0.132 & DASH & 0.084 & \#baseball & 0.164\\
\rowcolor{lightGray} row & 0.068 & not & 0.038 & 5 & 0.101 & PREP & 0.089 & CERTAIN & 0.143 & me & 0.130 & I & 0.077 & fan & 0.163 \\
got & 0.067 & why & 0.037 & SIXLTR & 0.100 & INGEST & 0.085 & TONE & 0.140 & look & 0.122 & DISCREP & 0.071 & MALE & 0.160 \\
\hline
\rowcolor{darkGray} \multicolumn{16}{|l|}{\bf POS (Unigrams and Bigrams)} \\
\hline
PRP\_VBD & 0.104 & NNP & 0.081 & CD\_NNS & 0.198 & DT\_NNP & 0.139 & RB\_JJ & 0.183 & VBP & 0.252 & \$\_CD & 0.161 & FW\_, & 0.164 \\
\rowcolor{lightGray} VBD & 0.093 & VB & 0.061 & VBD & 0.171 & VBP\_TO & 0.124 & VBP\_IN & 0.174 & PRP & 0.193 & \$ & 0.130 & VB\_VBD & 0.161 \\
CD\_NNS & 0.077 & RB\_VB & 0.056 & CD & 0.164 & IN\_: & 0.117 & VB\_RBR & 0.161 & PRP\_VBP & 0.191 & NN\_PDT & 0.130 & CC\_UH & 0.159 \\
\rowcolor{lightGray} PRP\$ & 0.074 & NNP\_NNP & 0.049 & NNS & 0.145 & VBP\_WP & 0.116 & JJR\_WRB & 0.161 & VBP\_JJ & 0.162 & NNS\_UH & 0.122 & VBZ\_DT & 0.151\\
VBD\_DT & 0.062 & VBP\_PRP & 0.048 & VBD\_DT & 0.141 & NNP\_UH & 0.116 & RB\_VBZ & 0.146 & UH\_DT & 0.150 & SYM\_: & 0.114 & DT\_RBS & 0.146 \\
\rowcolor{lightGray} NN\_IN & 0.061 & VBZ & 0.039 & PRP\_VBD & 0.132 & NFP\_NNP & 0.116 & CC\_JJ & 0.143 & VBP\_DT & 0.150 & VBZ\_JJ & 0.110 & UH\_NNP & 0.145 \\
IN\_CD & 0.060 & MD & 0.035 & NN\_IN & 0.132 & NNP & 0.116 & VBD\_: & 0.131 & RB\_VB & 0.149 & VB\_PRP\$ & 0.109 & .\_SYM & 0.138 \\
\rowcolor{lightGray} IN\_PRP\$ & 0.060 & NNP\_VBZ & 0.033 & IN\_CD & 0.130 & NNP\_NNS & 0.114 & .\_VBG & 0.123 & MD & 0.149 & PRP\$\_JJ & 0.109 & NFP\_CC & 0.137 \\
PRP\$\_NN & 0.058 & RB\_RB & 0.031 & VBN & 0.129 & TO\_VB & 0.109 & UH\_WP & 0.118 & MD\_VB & 0.134 & .\_VBD & 0.109 & PRP\_PRP\$ & 0.136 \\
\rowcolor{lightGray} VBD\_PRP\$ & 0.057 & MD\_PRP & 0.031 & VB\_JJR & 0.109 & TO & 0.107 & POS\_RB & 0.118 & CC\_WP & 0.131 & NN\_PRP\$ & 0.106 & NN\_NN & 0.135 \\

\hline
\end{tabular}}
\caption{Feature correlations including unigrams (lowercase), LIWC (uppercase), part-of-speech (POS) unigrams and bigrams with bragging and non-bragging tweets (left) and bragging tweets grouped in six types (right), sorted by Pearson correlation (r). All correlations are significant at $p<.01$, two-tailed t-test.}
\label{t:unigram_liwc}
\end{table*}

\section{Analysis}

\paragraph{Linguistic Feature Analysis}

We analyze the linguistic features i.e. unigrams, LIWC and part-of speech (POS) tags associated with bragging and its types in all tweets of our data set. For this purpose, we first tag all tweets using the Twitter POS Tagger \citep{derczynski2013twitter}. Each tweet is represented as a bag-of-words distribution over POS unigrams and bigrams to reveal distinctive syntactic patterns of bragging and their types. For each unigram, LIWC and POS feature, we compute correlations between its distribution across posts and the label of the post. Then, we use the method introduced by~\citet{schwartz2013personality} to rank the features using univariate Pearson correlation with words normalized to sum up to unit for each tweet.

Table \ref{t:unigram_liwc} (left) presents the top 15 features from unigrams (lowercase) and LIWC (uppercase) and top 10 features from POS unigrams and bigrams correlated with bragging and non-bragging tweets. We notice that the top words in the bragging category can be classified into (a) personal pronouns (e.g. \textit{my}, \textit{I}) that usually indicate the author of the bragging statement; (b) words related to time (e.g. \textit{FOCUSPAST}, \textit{TIME}, \textit{during}); and (c) words related to a specific bragging target (e.g. \textit{RELATIV}, \textit{ACHIEVE}, \textit{REWARD}, \textit{managed}). These findings are in line with the indicators of positive self-disclosure  by \citet{dayter2018self} and \citet{bazarova2013managing}. Furthermore, personal pronouns followed by a verb in past tense (\textit{PRP\_VBD}) is common in bragging (e.g. \textit{I forgot what it's like to be good at school. Today I finished a thing we were doing so fast that everyone around me started asking ME for help instead of the prof :')})  

Table \ref{t:unigram_liwc} (right) presents the top 15 features from unigrams (lowercase) and LIWC (uppercase) correlated with bragging tweets grouped in six types. We observe that \textbf{Achievement} statements usually involve verbs that are in past tense or indicate a result (e.g. \textit{FOCUSPAST}, \textit{finished}, \textit{beat}). A POS pattern common in \textbf{Achievement} statements is a cardinal number followed by nouns in plural (\textit{CD\_NNS}), similar to its unigram and LIWC features (\textit{NUMBER}, \textit{3}, \textit{5}) (e.g. \textit{I made a total of 5 dollars from online surveys wooo}). It is worth noting that one of the prevalent LIWC features for \textbf{Action} is \textit{FOCUSFUTURE}. This is because the user may brag about a planned action (e.g. \textit{@USER You know what? I'm going to make some PizzaRolls Brag}). Most of the top words in \textbf{Feeling} express emotion or sensitivity (e.g. \textit{happy}, \textit{blessed}), which is consistent with the top POS feature, \textit{RB\_JJ} (e.g. \textit{absolutely chuffed}, \textit{so happy}). In \textbf{Trait} category, words are mostly pronouns (e.g. \textit{I}, \textit{PRP}, \textit{PRP\_VBP}) and verbs (e.g. \textit{VBP}, \textit{VBP\_JJ}). Words appear frequently in \textbf{Possession} category are actions related to purchase (e.g. \textit{own}, \textit{buy}) and nouns related to a tangible object (e.g. \textit{car}, \textit{bedroom}). In addition, users usually show off the value of their possessions using statements that involve currency signs (\textit{\$}) or currency signs followed by a number (\textit{\$\_CD}) (e.g. \textit{I just signed a new three-year contract and I'll be getting 235 anytime minutes per month. Plus, the company is going to throw in a phone for just \$ 49 per month. I'll bet you can't beat that deal!}). Finally, top words in \textbf{Affiliation} category involve positive feeling towards belonging to a group (e.g. \textit{proud}, \textit{amazing}) and nouns related to it (e.g. \textit{FAMILY}, \textit{team}).

\renewcommand{\arraystretch}{1.0}
\begin{table}[!t]
\small
\center
\begin{tabular}{|l|c|c|}
\hline
\rowcolor{darkGray} \bf Class & \bf Mean & \bf Median \\
\hline
Achievement & 3.06 & 3.00 \\
\rowcolor{lightGray} Action & 0.91 & 0 \\
Feeling & 0.50 & 0\\
\rowcolor{lightGray} Trait & 2.38 & 2.00 \\
Possession & 2.00 & 0.50\\
\rowcolor{lightGray} Affiliation & 5.50 & 2.00 \\
\hline
\end{tabular}
\caption{Mean and median Twitter favorites across bragging classes on a sample set of the data.}
\label{t:favorite}
\end{table}

\paragraph{Bragging and Post Popularity} We also analyze the association between bragging posts and the number of favorites/retweets they receive by other users. 
Similar to the previous linguistic feature analysis, we use univariate Pearson correlation to compute the correlations between the log-scaled favorites/retweets number of each tweet and its label (i.e. bragging or non-bragging) by controlling the numbers of followers and friends of the user who post the tweet. Our results show that the number of favorites is positively correlated with bragging (see Appendix Figure \ref{fig:correlation}) while there is no correlation between bragging and the number of retweets.

We further explore the popularity of different bragging types. We randomly analyze a set of 443 tweets containing 56 bragging statements, where the follower and friend number of users are within a similar range: from 100 to 500 followers and from 500 to 1000 friends ($r$ = 0.19, $p$ < .01). We compute the mean and median Twitter favorites across the six bragging classes (see Table \ref{t:favorite}). We observe that bragging statements about \textit{Affiliation} such as family members or sports teams are more likely to receive considerable amount of favorites with the mean of 5.5. For example, 14 users favorite the tweet \textit{This maybe is a little, but I'm SO proud of my research group. We represent so many different personality types, cultures, ways of thinking, etc, and every single member of my lab (all 21 of them)}. We speculate this is because praising the group that one belongs to instead of oneself as a bragging strategy enables users be perceived as more likeable. Furthermore, bragging about \textit{Achievement} is generally marked as favorite by other users with the median of 3, where bigger achievements in the content such as job offers may receive more favorites (e.g. tweet \textit{Scored 80 \% on my thesis. Rather proud of that given the circumstances: new baby; pandemic; late topic change due to lockdown; minimal uni support because of furloughs; and an international move.} was marked as favorite 15 times).

\paragraph{Class Confusion Analysis}

\begin{figure*}[t!]
\centering
\begin{minipage}{.47\textwidth}
    \center
    \includegraphics[scale=0.35]{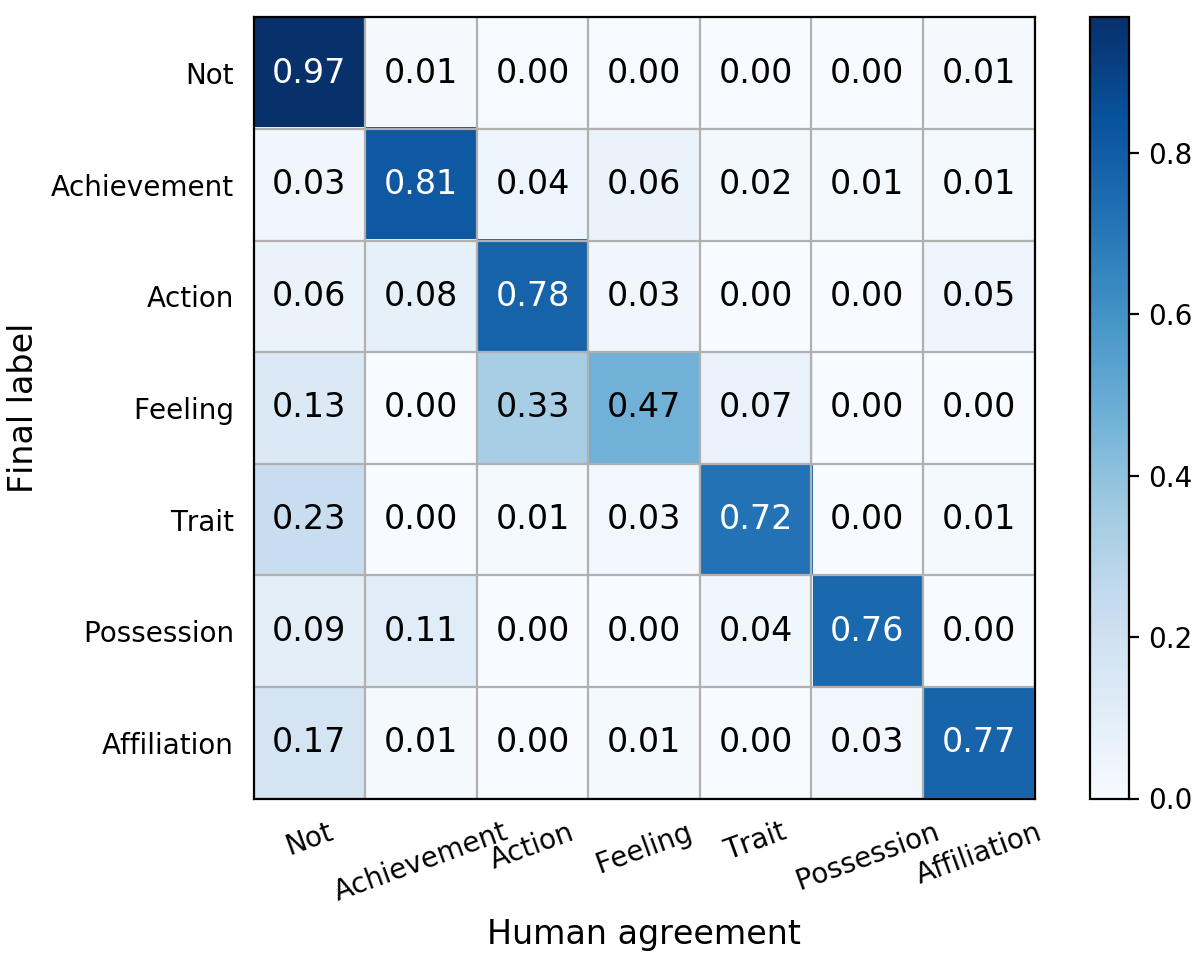}
    \caption{Confusion matrix of annotator agreement on seven bragging categories.}
    \label{fig:annotator_cm7} 
\end{minipage}
\quad
\begin{minipage}{.47\textwidth}
    \center
    \includegraphics[scale=0.35]{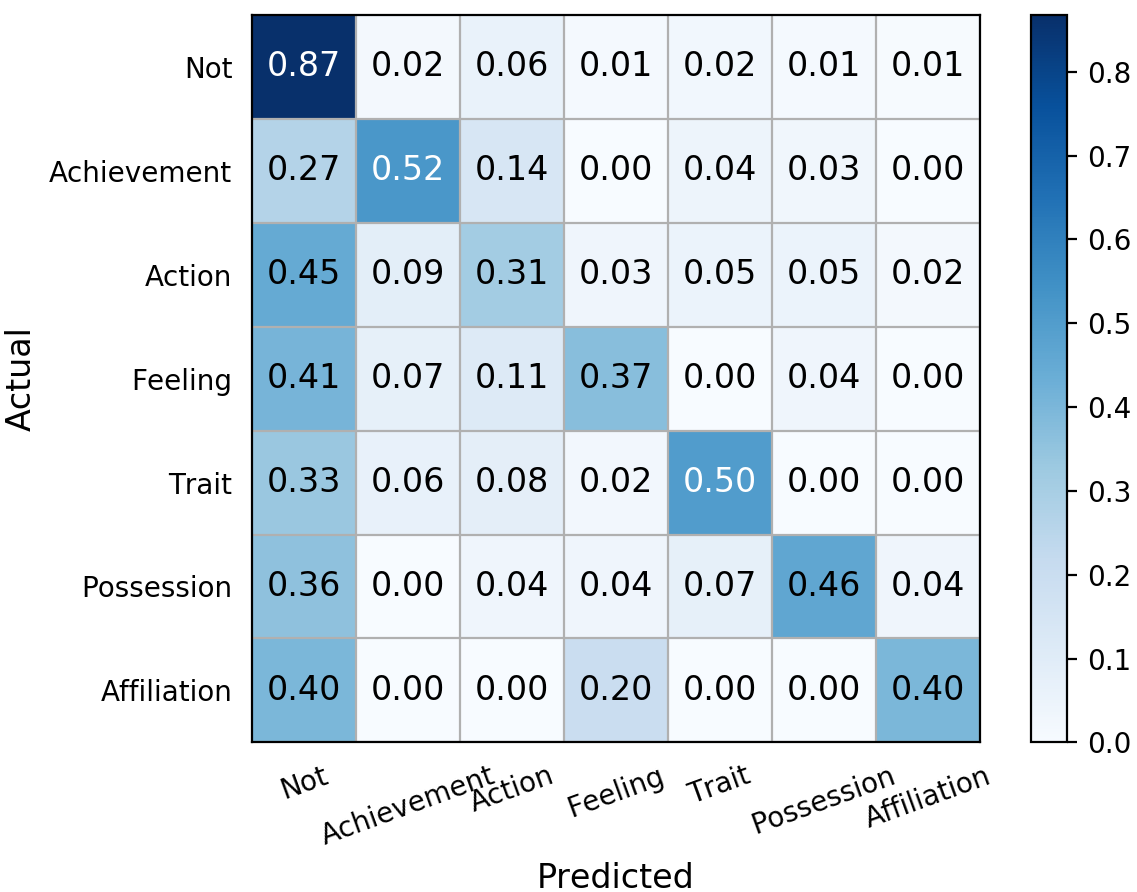}
    \caption{Confusion matrix of the best performing model on multi-class bragging classification, i.e. BERTweet-Clusters} 
    \label{fig:model_cm7} 
\end{minipage}
\end{figure*}


Figure \ref{fig:annotator_cm7} presents the confusion matrix of human agreement on seven classes normalized over the actual values (rows). We observe that \textit{Non-bragging} (97\%), \textit{Achievement} (81\%) and \textit{Action} (78\%) have high agreement, consistent with the class frequency. \textit{Affiliation} (77\%), \textit{Possession} (76\%) and \textit{Trait} (72\%) have comparable percentages as these are easily associated with a bragging target or group. The \textit{Feeling} category has the lowest percentage mostly caused by misclassification to the \textit{Action} category. This is due to the fact that both types are not associated to a concrete outcome by definition, with the \textit{feeling} class linked to a feeling linked to an action. Thus, it makes the boundary between bragging about the action or the feeling associated to the action more challenging to interpret. The next most frequent confusion is between \textit{possession} and \textit{achievement}, which usually arises when a tangible possession is involved and the annotators disagree if the author was bragging about the actual possession or the action that lead to the author obtaining that possession (e.g. \textit{@USER I just got some stealth 300 easily the best headset I've ever had going from astro to turtle beach was a night and day difference}).

Figure \ref{fig:model_cm7} presents the confusion matrix between bragging type predictions from the best performing model, BERTweet-Clusters, on the multi-class classification task. First, we observe that the model is more likely to misclassify other classes as the dominant class, \textit{Non-bragging}. Secondly, the most unambiguous classes are \textit{Non-bragging} (87\%) and \textit{Achievement} (52\%), which are in line with human agreement. Also, the model is good at identifying \textit{Trait} (50\%) and \textit{Possession} (46\%) due to the particular bragging targets (e.g. personalities, skills or tangible objects). Furthermore, we notice that the percentages of \textit{Action} (31\%) and \textit{Feeling} (37\%) are low. We speculate this is because they share more similarities with other classes (e.g. involving actions). This might also  explain the high percentage of misclassified data points between \textit{Action} and \textit{Achievement}, \textit{Feeling} and \textit{Action}. Lastly, the model often confuses \textit{Affiliation} with \textit{Feeling} likely because the terms that express positive feelings (e.g. `proud', \blueheart{}{}) also appear frequently in \textit{Affiliation} (see Table \ref{t:unigram_liwc}).


\paragraph{Error Analysis}

Finally, we perform an error analysis to examine the behavior and limitations of our best performing model (i.e. \textit{BERTweet-LIWC} for binary classification and \textit{BERTweet-Clusters} for multi-class classification) and identify pathways to improve the task modeling. 

We first start with the binary bragging classification. We observe that non-bragging tweets containing positive sentiment are easy to be misclassified as bragging and even if such tweets involve something valued positively by authors, the purpose is usually to express recommendation, compliment or appreciation to others: 
%
\begin{quote}
\small
    T1: \emph{@USER paid for my {\bf new bottle of vodka} \& {\bf I Love Her with all my heart \redheart{}{}}}
\end{quote}
Another frequent error happens when non-bragging tweets contain popular bragging targets such as achievement-oriented (e.g. weight loss, marathon) or possession-oriented (e.g. car, electronics):
\begin{quote}
\small
    T2: \emph{4 spaces left on my budget {\bf weight loss} program. £ 5 a week!???}
\end{quote}
Bragging often involves contextual understanding that goes beyond word use and require deep understanding of the context to determine the label. For example, common terms such as \textit{first}, \textit{finally}, \textit{just} often appear in both non-bragging (T3) and bragging (T4) tweets: 
\begin{quote}
\small
    T3: \emph{{\bf just cleaned} my cats' toilets} \\
    T4: \emph{It happened again! I {\bf just completed} 30 minutes of meditation with @USER. {\bf Just} sitting and resting in presence.}
\end{quote}

Models also fail to detect bragging mainly because it is indirect or there are no typical trigger terms, so they lean on pre-training to contextualize: 
\begin{quote}
\small
    T5: \emph{9 hr drives feel like nothing now lol}
\end{quote}

Some bragging statements use additional mitigation strategies, e.g. re-framing the bragging statement as irony, as a complaint or invoking praise from a third party: 
\begin{quote}
\small
    T6: \emph{I find it strange how I was always the weird one in school and irl but online people think im cool for some reason}
\end{quote}

Finally, we highlight some representative examples of model confusion between bragging types. One example is when users' actions lead or not to a concrete result. In this example the model predicted \textit{Action}, but the actual label is \textit{Achievement}:
\begin{quote}
\small
    T7: \emph{not to appropriate the gang escapes culture but me n my parents just did an escape room n actually got out?}
\end{quote}

Another example is an \textit{Action} misclassified as \textit{Possession}. This usually happens when a common phrase indicative of a certain type of bragging (\textit{a new dish)}) is invoked as part of an action:
\begin{quote}
\small
    T8: \emph{I had \textbf{a new dish} "egusi" it’s so damn good! Love Nigerian food!}
\end{quote}

Other errors occur when multiple types of bragging are present (e.g. feeling and action) but the label expresses the more salient type, such as the \textit{feeling} highlighted in this example:
\begin{quote}
\small
    T9: \emph{Literally \textbf{had the best time} with the girls last night, don’t think I’ve drank that much in my life?}
\end{quote}



\section{Conclusion}

We presented the first computational approach to analyzing and modeling bragging as a speech act along with its types in social media. We introduced a publicly available annotated data set in English  collected from Twitter. We experimented using transformer models combined with linguistic information on binary bragging and multiclass bragging type prediction. Finally, we presented an extensive analysis of features related to bragging statements and an error analysis of the model predictive behavior. In future work, we 
plan to study the extent to which bragging is used across various locations~\cite{sanchez-villegas-etal-2020-point,sanchez-villegas-aletras-2021-point} and languages and how it is employed by users across contexts. 

\section*{Acknowledgements}

We would like to thank Ari Silburt, Danae S\'{a}nchez Villegas, Yida Mu, and all the anonymous reviewers for their valuable feedback.

\section*{Ethics Statement}
Our work has received approval from the Ethics Committee of the Department of Computer Science at the University of Sheffield (No 037572) and complies with Twitter's data policy for research.\footnote{https://developer.twitter.com/en/developer-terms/agreement-and-policy} 



\bibliography{anthology,custom}

\begin{thebibliography}{81}
\expandafter\ifx\csname natexlab\endcsname\relax\def\natexlab#1{#1}\fi

\bibitem[{Artstein and Poesio(2008)}]{artstein2008inter}
Ron Artstein and Massimo Poesio. 2008.
\newblock {Inter-coder agreement for Computational Linguistics}.
\newblock \emph{Computational Linguistics}, 34(4):555--596.

\bibitem[{Bak et~al.(2012)Bak, Kim, and Oh}]{bak-etal-2012-self}
JinYeong Bak, Suin Kim, and Alice Oh. 2012.
\newblock \href {https://aclanthology.org/P12-2012} {Self-disclosure and
  relationship strength in {T}witter conversations}.
\newblock In \emph{Proceedings of the 50th Annual Meeting of the Association
  for Computational Linguistics (Volume 2: Short Papers)}, pages 60--64, Jeju
  Island, Korea. Association for Computational Linguistics.

\bibitem[{Bak et~al.(2014{\natexlab{a}})Bak, Lin, and Oh}]{bak2014self}
JinYeong Bak, Chin-Yew Lin, and Alice Oh. 2014{\natexlab{a}}.
\newblock \href {https://www.aclweb.org/anthology/D14-1213} {Self-disclosure
  topic model for classifying and analyzing {T}witter conversations}.
\newblock In \emph{Proceedings of the 2014 Conference on Empirical Methods in
  Natural Language Processing (EMNLP)}, pages 1986--1996.

\bibitem[{Bak et~al.(2014{\natexlab{b}})Bak, Lin, and Oh}]{bak2014sdtm}
JinYeong Bak, Chin-Yew Lin, and Alice Oh. 2014{\natexlab{b}}.
\newblock \href {https://doi.org/10.3115/v1/W14-2706} {Self-disclosure topic
  model for {T}witter conversations}.
\newblock In \emph{Proceedings of the Joint Workshop on Social Dynamics and
  Personal Attributes in Social Media}, pages 42--49, Baltimore, Maryland.
  Association for Computational Linguistics.

\bibitem[{Bamman and Smith(2015)}]{bamman2015contextualized}
David Bamman and Noah~A Smith. 2015.
\newblock {Contextualized Sarcasm Detection on Twitter}.
\newblock In \emph{Proceedings of the 9th International Conference on Weblogs
  and Social Media}, ICWSM, pages 574--577.

\bibitem[{Baumeister(1982)}]{baumeister1982self}
Roy~F Baumeister. 1982.
\newblock A self-presentational view of social phenomena.
\newblock \emph{Psychological bulletin}, 91(1):3.

\bibitem[{Bazarova et~al.(2013)Bazarova, Taft, Choi, and
  Cosley}]{bazarova2013managing}
Natalya~N Bazarova, Jessie~G Taft, Yoon~Hyung Choi, and Dan Cosley. 2013.
\newblock \href {https://doi.org/10.1177/0261927X12456384} {Managing
  {I}mpressions and {R}elationships on {F}acebook: Self-{P}resentational and
  {R}elational {C}oncerns {R}evealed {T}hrough the {A}nalysis of {L}anguage
  {S}tyle}.
\newblock \emph{Journal of Language and Social Psychology}, 32(2):121--141.

\bibitem[{Beck et~al.(2021)Beck, Lee, Viehmann, Maurer, Quiring, and
  Gurevych}]{beck-etal-2021-investigating}
Tilman Beck, Ji-Ung Lee, Christina Viehmann, Marcus Maurer, Oliver Quiring, and
  Iryna Gurevych. 2021.
\newblock \href {https://doi.org/10.18653/v1/2021.acl-long.1} {Investigating
  label suggestions for opinion mining in {G}erman covid-19 social media}.
\newblock In \emph{Proceedings of the 59th Annual Meeting of the Association
  for Computational Linguistics and the 11th International Joint Conference on
  Natural Language Processing (Volume 1: Long Papers)}, pages 1--13, Online.
  Association for Computational Linguistics.

\bibitem[{Brown and Levinson(1987)}]{brown1987politeness}
Penelope Brown and Stephen~C Levinson. 1987.
\newblock \emph{Politeness: Some universals in language usage}, volume~4.
\newblock Cambridge University Press.

\bibitem[{Chen et~al.(2016)Chen, Li, Hu, and Li}]{chen2016anonymity}
Xi~Chen, Gang Li, YunDi Hu, and Yujie Li. 2016.
\newblock How {A}nonymity {I}nfluence {S}elf-disclosure {T}endency on {S}ina
  {W}eibo: An {E}mpirical {S}tudy.
\newblock \emph{The anthropologist}, 26(3):217--226.

\bibitem[{Chen et~al.(2020)Chen, Levitan, Levine, Mandic, and
  Hirschberg}]{chen-etal-2020-acoustic}
Xi~(Leslie) Chen, Sarah~Ita Levitan, Michelle Levine, Marko Mandic, and Julia
  Hirschberg. 2020.
\newblock \href {https://doi.org/10.1162/tacl_a_00311} {Acoustic-prosodic and
  lexical cues to deception and trust: Deciphering how people detect lies}.
\newblock \emph{Transactions of the Association for Computational Linguistics},
  8:199--214.

\bibitem[{Chinnappa and Blanco(2018)}]{chinnappa-blanco-2018-mining}
Dhivya Chinnappa and Eduardo Blanco. 2018.
\newblock \href {https://doi.org/10.18653/v1/N18-1046} {Mining possessions:
  Existence, type and temporal anchors}.
\newblock In \emph{Proceedings of the 2018 Conference of the North {A}merican
  Chapter of the Association for Computational Linguistics: Human Language
  Technologies, Volume 1 (Long Papers)}, pages 496--505, New Orleans,
  Louisiana. Association for Computational Linguistics.

\bibitem[{Cho et~al.(2014)Cho, van Merri{\"e}nboer, Gulcehre, Bahdanau,
  Bougares, Schwenk, and Bengio}]{cho-etal-2014-learning}
Kyunghyun Cho, Bart van Merri{\"e}nboer, Caglar Gulcehre, Dzmitry Bahdanau,
  Fethi Bougares, Holger Schwenk, and Yoshua Bengio. 2014.
\newblock \href {https://doi.org/10.3115/v1/D14-1179} {Learning phrase
  representations using {RNN} encoder{--}decoder for statistical machine
  translation}.
\newblock In \emph{Proceedings of the 2014 Conference on Empirical Methods in
  Natural Language Processing ({EMNLP})}, pages 1724--1734, Doha, Qatar.
  Association for Computational Linguistics.

\bibitem[{Chou and Edge(2012)}]{chou2012they}
Hui-Tzu~Grace Chou and Nicholas Edge. 2012.
\newblock “they are happier and having better lives than i am”: The impact
  of using facebook on perceptions of others' lives.
\newblock \emph{Cyberpsychology, behavior, and social networking},
  15(2):117--121.

\bibitem[{Chung et~al.(2019)Chung, Kuzmenko, Tekiroglu, and
  Guerini}]{chung-etal-2019-conan}
Yi-Ling Chung, Elizaveta Kuzmenko, Serra~Sinem Tekiroglu, and Marco Guerini.
  2019.
\newblock \href {https://doi.org/10.18653/v1/P19-1271} {{CONAN} - {CO}unter
  {NA}rratives through nichesourcing: a multilingual dataset of responses to
  fight online hate speech}.
\newblock In \emph{Proceedings of the 57th Annual Meeting of the Association
  for Computational Linguistics}, pages 2819--2829, Florence, Italy.
  Association for Computational Linguistics.

\bibitem[{Danescu-Niculescu-Mizil et~al.(2013)Danescu-Niculescu-Mizil, Sudhof,
  Jurafsky, Leskovec, and
  Potts}]{danescu-niculescu-mizil-etal-2013-computational}
Cristian Danescu-Niculescu-Mizil, Moritz Sudhof, Dan Jurafsky, Jure Leskovec,
  and Christopher Potts. 2013.
\newblock \href {https://aclanthology.org/P13-1025} {A computational approach
  to politeness with application to social factors}.
\newblock In \emph{Proceedings of the 51st Annual Meeting of the Association
  for Computational Linguistics (Volume 1: Long Papers)}, pages 250--259,
  Sofia, Bulgaria. Association for Computational Linguistics.

\bibitem[{Dayter(2014)}]{dayter2014self}
Daria Dayter. 2014.
\newblock \href {https://doi.org/10.1016/j.pragma.2013.11.021} {Self-praise in
  microblogging}.
\newblock \emph{Journal of Pragmatics}, 61:91--102.

\bibitem[{Dayter(2018)}]{dayter2018self}
Daria Dayter. 2018.
\newblock Self-praise online and offline: The hallmark speech act of social
  media?
\newblock \emph{Internet Pragmatics}, 1(1):184--203.

\bibitem[{Derczynski et~al.(2013)Derczynski, Ritter, Clark, and
  Bontcheva}]{derczynski2013twitter}
Leon Derczynski, Alan Ritter, Sam Clark, and Kalina Bontcheva. 2013.
\newblock Twitter {P}art-of-{S}peech {T}agging for {A}ll: Overcoming {S}parse
  and {N}oisy {D}ata.
\newblock In \emph{Proceedings of the international conference recent advances
  in natural language processing ranlp 2013}, pages 198--206.

\bibitem[{Devlin et~al.(2019)Devlin, Chang, Lee, and
  Toutanova}]{devlin-etal-2019-bert}
Jacob Devlin, Ming-Wei Chang, Kenton Lee, and Kristina Toutanova. 2019.
\newblock \href {https://doi.org/10.18653/v1/N19-1423} {{BERT}: Pre-training of
  deep bidirectional transformers for language understanding}.
\newblock In \emph{Proceedings of the 2019 Conference of the North {A}merican
  Chapter of the Association for Computational Linguistics: Human Language
  Technologies, Volume 1 (Long and Short Papers)}, pages 4171--4186,
  Minneapolis, Minnesota. Association for Computational Linguistics.

\bibitem[{Gilmore and Ferris(1989)}]{gilmore1989effects}
David~C Gilmore and Gerald~R Ferris. 1989.
\newblock The effects of applicant impression management tactics on interviewer
  judgments.
\newblock \emph{Journal of management}, 15(4):557--564.

\bibitem[{Goffman et~al.(1978)}]{goffman1978presentation}
Erving Goffman et~al. 1978.
\newblock \emph{The Presentation of Self in Everyday Life}, volume~21.
\newblock Harmondsworth London.

\bibitem[{Gu(1990)}]{gu1990politeness}
Yueguo Gu. 1990.
\newblock Politeness phenomena in modern chinese.
\newblock \emph{Journal of pragmatics}, 14(2):237--257.

\bibitem[{Halpern et~al.(2017)Halpern, Katz, and Carril}]{halpern2017online}
Daniel Halpern, James~E Katz, and Camila Carril. 2017.
\newblock The online ideal persona vs. the jealousy effect: Two explanations of
  why selfies are associated with lower-quality romantic relationships.
\newblock \emph{Telematics and Informatics}, 34(1):114--123.

\bibitem[{Herbert(1990)}]{herbert1990sex}
Robert~K Herbert. 1990.
\newblock Sex-based differences in compliment behavior1.
\newblock \emph{Language in society}, 19(2):201--224.

\bibitem[{Hogan(1982)}]{hogan1982socioanalytic}
Robert Hogan. 1982.
\newblock A socioanalytic theory of personality.
\newblock In \emph{Nebraska symposium on motivation}. University of Nebraska
  Press.

\bibitem[{Holtgraves(1990)}]{holtgraves1990language}
Thomas Holtgraves. 1990.
\newblock The language of self-disclosure.

\bibitem[{Hoorens et~al.(2012)Hoorens, Pandelaere, Oldersma, and
  Sedikides}]{hoorens2012hubris}
Vera Hoorens, Mario Pandelaere, Frans Oldersma, and Constantine Sedikides.
  2012.
\newblock \href {https://doi.org/10.1111/j.1467-6494.2011.00759.x} {The
  {H}ubris {H}ypothesis: You {C}an {S}elf-{E}nhance, {B}ut {Y}ou'd {B}etter
  {N}ot {S}how {I}t}.
\newblock \emph{Journal of personality}, 80(5):1237--1274.

\bibitem[{Jin and Aletras(2020)}]{jin-aletras-2020-complaint}
Mali Jin and Nikolaos Aletras. 2020.
\newblock \href {https://doi.org/10.18653/v1/2020.coling-main.157} {Complaint
  identification in social media with transformer networks}.
\newblock In \emph{Proceedings of the 28th International Conference on
  Computational Linguistics}, pages 1765--1771, Barcelona, Spain (Online).
  International Committee on Computational Linguistics.

\bibitem[{Jin and Aletras(2021)}]{jin-aletras-2021-modeling}
Mali Jin and Nikolaos Aletras. 2021.
\newblock \href {https://doi.org/10.18653/v1/2021.naacl-main.180} {Modeling the
  severity of complaints in social media}.
\newblock In \emph{Proceedings of the 2021 Conference of the North American
  Chapter of the Association for Computational Linguistics: Human Language
  Technologies}, pages 2264--2274, Online. Association for Computational
  Linguistics.

\bibitem[{Jones(1990)}]{jones1990interpersonal}
Edward~E Jones. 1990.
\newblock \emph{Interpersonal perception}.
\newblock WH Freeman/Times Books/Henry Holt \& Co.

\bibitem[{Jones et~al.(1982)Jones, Pittman et~al.}]{jones1982toward}
Edward~E Jones, Thane~S Pittman, et~al. 1982.
\newblock Toward a general theory of strategic self-presentation.
\newblock \emph{Psychological perspectives on the self}, 1(1):231--262.

\bibitem[{Kingma and Adam(2015)}]{kingma2015method}
Diederik~P Kingma and Jimmy~Ba Adam. 2015.
\newblock Adam: A {M}ethod for {S}tochastic {O}ptimization.
\newblock \emph{Optimization. In, ICLR}, 5.

\bibitem[{Krippendorff(2011)}]{krippendorff2011computing}
Klaus Krippendorff. 2011.
\newblock Computing {K}rippendorff's {A}lpha-{R}eliability.

\bibitem[{Lampos et~al.(2014)Lampos, Aletras, Preo{\c{t}}iuc-Pietro, and
  Cohn}]{lampos-etal-2014-predicting}
Vasileios Lampos, Nikolaos Aletras, Daniel Preo{\c{t}}iuc-Pietro, and Trevor
  Cohn. 2014.
\newblock \href {https://doi.org/10.3115/v1/E14-1043} {Predicting and
  characterising user impact on {T}witter}.
\newblock In \emph{Proceedings of the 14th Conference of the {E}uropean Chapter
  of the Association for Computational Linguistics}, pages 405--413,
  Gothenburg, Sweden. Association for Computational Linguistics.

\bibitem[{Leary and Kowalski(1990)}]{leary1990impression}
Mark~R Leary and Robin~M Kowalski. 1990.
\newblock Impression management: A literature review and two-component model.
\newblock \emph{Psychological bulletin}, 107(1):34.

\bibitem[{Lee-Won et~al.(2014)Lee-Won, Shim, Joo, and Park}]{lee2014puts}
Roselyn~J Lee-Won, Minsun Shim, Yeon~Kyoung Joo, and Sung~Gwan Park. 2014.
\newblock Who puts the best “face” forward on facebook?: Positive
  self-presentation in online social networking and the role of
  self-consciousness, actual-to-total friends ratio, and culture.
\newblock \emph{Computers in Human Behavior}, 39:413--423.

\bibitem[{Leech(2016)}]{leech2016principles}
Geoffrey Leech. 2016.
\newblock \emph{Principles of pragmatics}.
\newblock Routledge.

\bibitem[{Levontin and Yom-Tov(2017)}]{levontin2017negative}
Liat Levontin and Elad Yom-Tov. 2017.
\newblock Negative self-disclosure on the web: the role of guilt relief.
\newblock \emph{Frontiers in psychology}, 8:1068.

\bibitem[{Liu et~al.(2019)Liu, Ott, Goyal, Du, Joshi, Chen, Levy, Lewis,
  Zettlemoyer, and Stoyanov}]{liu2019roberta}
Yinhan Liu, Myle Ott, Naman Goyal, Jingfei Du, Mandar Joshi, Danqi Chen, Omer
  Levy, Mike Lewis, Luke Zettlemoyer, and Veselin Stoyanov. 2019.
\newblock {RoBERTa}: A {R}obustly {O}ptimized {BERT} {P}retraining {A}pproach.
\newblock \emph{arXiv preprint arXiv:1907.11692}.

\bibitem[{Maronikolakis et~al.(2020)Maronikolakis, S{\'a}nchez~Villegas,
  Preotiuc-Pietro, and Aletras}]{maronikolakis-etal-2020-analyzing}
Antonis Maronikolakis, Danae S{\'a}nchez~Villegas, Daniel Preotiuc-Pietro, and
  Nikolaos Aletras. 2020.
\newblock \href {https://doi.org/10.18653/v1/2020.acl-main.403} {Analyzing
  political parody in social media}.
\newblock In \emph{Proceedings of the 58th Annual Meeting of the Association
  for Computational Linguistics}, pages 4373--4384, Online. Association for
  Computational Linguistics.

\bibitem[{Matley(2018)}]{matley2018not}
David Matley. 2018.
\newblock \href {https://doi.org/10.1016/j.dcm.2017.07.007} {“this is {N}ot
  a\# humblebrag, this is just a\# brag”: The pragmatics of self-praise,
  hashtags and politeness in {I}nstagram posts}.
\newblock \emph{Discourse, context \& media}, 22:30--38.

\bibitem[{Matley(2020)}]{matley2020isn}
David Matley. 2020.
\newblock Isn’t working on the weekend the worst?\# humblebrag”: The impact
  of irony and hashtag use on the perception of self-praise in instagram posts.

\bibitem[{Mendelsohn et~al.(2021)Mendelsohn, Budak, and
  Jurgens}]{mendelsohn-etal-2021-modeling}
Julia Mendelsohn, Ceren Budak, and David Jurgens. 2021.
\newblock \href {https://doi.org/10.18653/v1/2021.naacl-main.179} {Modeling
  framing in immigration discourse on social media}.
\newblock In \emph{Proceedings of the 2021 Conference of the North American
  Chapter of the Association for Computational Linguistics: Human Language
  Technologies}, pages 2219--2263, Online. Association for Computational
  Linguistics.

\bibitem[{Michikyan et~al.(2015)Michikyan, Dennis, and
  Subrahmanyam}]{michikyan2015can}
Minas Michikyan, Jessica Dennis, and Kaveri Subrahmanyam. 2015.
\newblock Can {Y}ou {G}uess {W}ho {I} {A}m? {R}eal, {I}deal, and {F}alse
  {S}elf-{P}resentation on {F}acebook among {E}merging {A}dults.
\newblock \emph{Emerging Adulthood}, 3(1):55--64.

\bibitem[{Miller et~al.(1992)Miller, Lee~Cooke, Tsang, and
  Morgan}]{miller1992should}
Lynn~Carol Miller, Linda Lee~Cooke, Jennifer Tsang, and Faith Morgan. 1992.
\newblock Should {I} {B}rag? {N}ature and {I}mpact of {P}ositive and {B}oastful
  {D}isclosures for {W}omen and {M}en.
\newblock \emph{Human Communication Research}, 18(3):364--399.

\bibitem[{Mohammad et~al.(2016)Mohammad, Kiritchenko, Sobhani, Zhu, and
  Cherry}]{mohammad2016semeval}
Saif Mohammad, Svetlana Kiritchenko, Parinaz Sobhani, Xiaodan Zhu, and Colin
  Cherry. 2016.
\newblock {Semeval-2016 task 6: Detecting Stance in Tweets}.
\newblock In \emph{Proceedings of the 10th International Workshop on Semantic
  Evaluation (SemEval-2016)}, *SEM, pages 31--41.

\bibitem[{Mohammad et~al.(2018)Mohammad, Bravo-Marquez, Salameh, and
  Kiritchenko}]{SemEval2018Task1}
Saif~M. Mohammad, Felipe Bravo-Marquez, Mohammad Salameh, and Svetlana
  Kiritchenko. 2018.
\newblock Semeval-2018 {T}ask 1: {A}ffect in tweets.
\newblock In \emph{Proceedings of International Workshop on Semantic Evaluation
  (SemEval-2018)}, *SEM, pages 1--17.

\bibitem[{Mohammad and Turney(2013)}]{mohammad2013crowdsourcing}
Saif~M Mohammad and Peter~D Turney. 2013.
\newblock Crowdsourcing a {W}ord--{E}motion {A}ssociation {L}exicon.
\newblock \emph{Computational intelligence}, 29(3):436--465.

\bibitem[{Nguyen et~al.(2020)Nguyen, Vu, and
  Tuan~Nguyen}]{nguyen-etal-2020-bertweet}
Dat~Quoc Nguyen, Thanh Vu, and Anh Tuan~Nguyen. 2020.
\newblock \href {https://doi.org/10.18653/v1/2020.emnlp-demos.2} {{BERT}weet: A
  pre-trained language model for {E}nglish tweets}.
\newblock In \emph{Proceedings of the 2020 Conference on Empirical Methods in
  Natural Language Processing: System Demonstrations}, pages 9--14, Online.
  Association for Computational Linguistics.

\bibitem[{Oprea and Magdy(2020)}]{oprea-magdy-2020-isarcasm}
Silviu Oprea and Walid Magdy. 2020.
\newblock \href {https://doi.org/10.18653/v1/2020.acl-main.118} {i{S}arcasm: A
  dataset of intended sarcasm}.
\newblock In \emph{Proceedings of the 58th Annual Meeting of the Association
  for Computational Linguistics}, pages 1279--1289, Online. Association for
  Computational Linguistics.

\bibitem[{Paramita and Septianto(2021)}]{paramita2021benefits}
Widya Paramita and Felix Septianto. 2021.
\newblock The benefits and pitfalls of humblebragging in social media
  advertising: the moderating role of the celebrity versus influencer.
\newblock \emph{International Journal of Advertising}, pages 1--24.

\bibitem[{Pennebaker et~al.(2001)Pennebaker, Francis, and
  Booth}]{pennebaker2001linguistic}
James~W Pennebaker, Martha~E Francis, and Roger~J Booth. 2001.
\newblock Linguistic {I}nquiry and {W}ord {C}ount: {LIWC} 2001.
\newblock \emph{Mahway: Lawrence Erlbaum Associates}, 71(2001):2001.

\bibitem[{Pennington et~al.(2014)Pennington, Socher, and
  Manning}]{pennington2014glove}
Jeffrey Pennington, Richard Socher, and Christopher~D Manning. 2014.
\newblock Glo{V}e: {G}lobal {V}ectors for {W}ord {R}epresentation.
\newblock In \emph{Proceedings of the 2014 conference on empirical methods in
  natural language processing (EMNLP)}, pages 1532--1543.

\bibitem[{Preo{\c{t}}iuc-Pietro et~al.(2019)Preo{\c{t}}iuc-Pietro, Gaman, and
  Aletras}]{preotiuc-pietro-etal-2019-automatically}
Daniel Preo{\c{t}}iuc-Pietro, Mihaela Gaman, and Nikolaos Aletras. 2019.
\newblock \href {https://doi.org/10.18653/v1/P19-1495} {Automatically
  identifying complaints in social media}.
\newblock In \emph{Proceedings of the 57th Annual Meeting of the Association
  for Computational Linguistics}, pages 5008--5019, Florence, Italy.
  Association for Computational Linguistics.

\bibitem[{Preo{\c{t}}iuc-Pietro et~al.(2015)Preo{\c{t}}iuc-Pietro, Volkova,
  Lampos, Bachrach, and Aletras}]{preoctiuc2015studying}
Daniel Preo{\c{t}}iuc-Pietro, Svitlana Volkova, Vasileios Lampos, Yoram
  Bachrach, and Nikolaos Aletras. 2015.
\newblock \href {https://doi.org/10.1371/journal.pone.0138717} {Studying {U}ser
  {I}ncome through {L}anguage, {B}ehaviour and {A}ffect in {S}ocial {M}edia}.
\newblock \emph{PloS one}, 10(9):e0138717.

\bibitem[{Rahman et~al.(2020)Rahman, Hasan, Lee, Zadeh, Mao, Morency, and
  Hoque}]{rahman2020integrating}
Wasifur Rahman, Md~Kamrul Hasan, Sangwu Lee, Amir Zadeh, Chengfeng Mao,
  Louis-Philippe Morency, and Ehsan Hoque. 2020.
\newblock \href {https://doi.org/10.18653/v1/2020.acl-main.214} {Integrating
  {M}ultimodal {I}nformation in {L}arge {P}retrained {T}ransformers}.
\newblock In \emph{Proceedings of the conference. Association for Computational
  Linguistics. Meeting}, volume 2020, page 2359. NIH Public Access.

\bibitem[{Ravichander and Black(2018)}]{ravichander-black-2018-empirical}
Abhilasha Ravichander and Alan~W. Black. 2018.
\newblock \href {https://doi.org/10.18653/v1/W18-5030} {An empirical study of
  self-disclosure in spoken dialogue systems}.
\newblock In \emph{Proceedings of the 19th Annual {SIG}dial Meeting on
  Discourse and Dialogue}, pages 253--263, Melbourne, Australia. Association
  for Computational Linguistics.

\bibitem[{Reinecke and Trepte(2014)}]{reinecke2014authenticity}
Leonard Reinecke and Sabine Trepte. 2014.
\newblock Authenticity and well-being on social network sites: A two-wave
  longitudinal study on the effects of online authenticity and the positivity
  bias in sns communication.
\newblock \emph{Computers in Human Behavior}, 30:95--102.

\bibitem[{Ren and Guo(2020)}]{ren2020self}
Wei Ren and Yaping Guo. 2020.
\newblock \href {https://doi.org/10.1016/j.pragma.2020.09.009} {Self-praise on
  {C}hinese social networking sites}.
\newblock \emph{Journal of Pragmatics}, 169:179--189.

\bibitem[{Rosenthal et~al.(2017)Rosenthal, Farra, and
  Nakov}]{rosenthal2017semeval}
Sara Rosenthal, Noura Farra, and Preslav Nakov. 2017.
\newblock {SemEval-2017 Task 4: Sentiment analysis in Twitter}.
\newblock In \emph{Proceedings of the 11th International Workshop on Semantic
  Evaluation (SemEval-2017)}, *SEM, pages 502--518.

\bibitem[{R{\"u}diger and Dayter(2020)}]{rudiger2020manbragging}
Sofia R{\"u}diger and Daria Dayter. 2020.
\newblock Manbragging online: Self-praise on pick-up artists’ forums.
\newblock \emph{Journal of Pragmatics}, 161:16--27.

\bibitem[{S{\'a}nchez~Villegas and
  Aletras(2021)}]{sanchez-villegas-aletras-2021-point}
Danae S{\'a}nchez~Villegas and Nikolaos Aletras. 2021.
\newblock \href {https://aclanthology.org/2021.emnlp-main.614}
  {Point-of-interest type prediction using text and images}.
\newblock In \emph{Proceedings of the 2021 Conference on Empirical Methods in
  Natural Language Processing}, pages 7785--7797, Online and Punta Cana,
  Dominican Republic. Association for Computational Linguistics.

\bibitem[{S{\'a}nchez~Villegas et~al.(2020)S{\'a}nchez~Villegas,
  Preotiuc-Pietro, and Aletras}]{sanchez-villegas-etal-2020-point}
Danae S{\'a}nchez~Villegas, Daniel Preotiuc-Pietro, and Nikolaos Aletras. 2020.
\newblock \href {https://aclanthology.org/2020.aacl-main.80} {Point-of-interest
  type inference from social media text}.
\newblock In \emph{Proceedings of the 1st Conference of the Asia-Pacific
  Chapter of the Association for Computational Linguistics and the 10th
  International Joint Conference on Natural Language Processing}, pages
  804--810, Suzhou, China. Association for Computational Linguistics.

\bibitem[{Schlenker(1980)}]{schlenker1980impression}
Barry~R Schlenker. 1980.
\newblock \emph{Impression management}, volume 222.

\bibitem[{Schwartz et~al.(2013)Schwartz, Eichstaedt, Kern, Dziurzynski,
  Ramones, Agrawal, Shah, Kosinski, Stillwell, and
  Seligman}]{schwartz2013personality}
H~Andrew Schwartz, Johannes~C Eichstaedt, Margaret~L Kern, Lukasz Dziurzynski,
  Stephanie~M Ramones, Megha Agrawal, Achal Shah, Michal Kosinski, David
  Stillwell, and Martin~EP Seligman. 2013.
\newblock {Personality, Gender, and Age in the Language of Social Media: The
  Open-vocabulary Approach}.
\newblock \emph{PloS ONE}, 8(9).

\bibitem[{Scopelliti et~al.(2015)Scopelliti, Loewenstein, and
  Vosgerau}]{scopelliti2015you}
Irene Scopelliti, George Loewenstein, and Joachim Vosgerau. 2015.
\newblock You call it “self-exuberance”; i call it “bragging”
  miscalibrated predictions of emotional responses to self-promotion.
\newblock \emph{Psychological science}, 26(6):903--914.

\bibitem[{Sedikides(1993)}]{sedikides1993assessment}
Constantine Sedikides. 1993.
\newblock Assessment, enhancement, and verification determinants of the
  self-evaluation process.
\newblock \emph{Journal of personality and social psychology}, 65(2):317.

\bibitem[{Sedikides et~al.(2007)Sedikides, Gregg, and Hart}]{sedikides2007}
Constantine Sedikides, Aiden~P. Gregg, and Claire~M. Hart. 2007.
\newblock The importance of being modest.
\newblock \emph{The Self: Frontiers in Social Psychology, edited by Constantine
  Sedikides and Stephen J. Spencer}, 163(84).

\bibitem[{Sezer et~al.(2018)Sezer, Gino, and Norton}]{sezer2018humblebragging}
Ovul Sezer, Francesca Gino, and Michael~I Norton. 2018.
\newblock Humblebragging: A distinct—and ineffective—self-presentation
  strategy.
\newblock \emph{Journal of Personality and Social Psychology}, 114(1):52.

\bibitem[{Tetlock(2002)}]{tetlock2002social}
Philip~E Tetlock. 2002.
\newblock Social functionalist frameworks for judgment and choice: intuitive
  politicians, theologians, and prosecutors.
\newblock \emph{Psychological review}, 109(3):451.

\bibitem[{Tian et~al.(2018)Tian, Rong, Shi, Liu, and Xiong}]{tian2018attention}
Zhengxi Tian, Wenge Rong, Libin Shi, Jingshuang Liu, and Zhang Xiong. 2018.
\newblock Attention {A}ware {B}idirectional {G}ated {R}ecurrent {U}nit {B}ased
  {F}ramework for {S}entiment {A}nalysis.
\newblock In \emph{International Conference on Knowledge Science, Engineering
  and Management}, pages 67--78. Springer.

\bibitem[{Tice et~al.(1995)Tice, Butler, Muraven, and
  Stillwell}]{tice1995modesty}
Dianne~M Tice, Jennifer~L Butler, Mark~B Muraven, and Arlene~M Stillwell. 1995.
\newblock \href {https://doi.org/10.1037/0022-3514.69.6.1120} {When modesty
  prevails: Differential favorability of self-presentation to friends and
  strangers.}
\newblock \emph{Journal of personality and social psychology}, 69(6):1120.

\bibitem[{Tobback(2019)}]{tobback2019telling}
Els Tobback. 2019.
\newblock \href {https://doi.org/10.1177/1750481319868854} {Telling the world
  how skilful you are: Self-praise strategies on {L}inkedin}.
\newblock \emph{Discourse \& Communication}, 13(6):647--668.

\bibitem[{Van~Damme et~al.(2017)Van~Damme, Deschrijver, Van~Geert, and
  Hoorens}]{van2017praising}
Carolien Van~Damme, Eliane Deschrijver, Eline Van~Geert, and Vera Hoorens.
  2017.
\newblock \href {https://doi.org/10.1177/0146167217703951} {When {P}raising
  {Y}ourself {I}nsults {O}thers: {S}elf-{S}uperiority {C}laims {P}rovoke
  {A}ggression}.
\newblock \emph{Personality and Social Psychology Bulletin}, 43(7):1008--1019.

\bibitem[{Vaswani et~al.(2017)Vaswani, Shazeer, Parmar, Uszkoreit, Jones,
  Gomez, Kaiser, and Polosukhin}]{vaswani2017attention}
Ashish Vaswani, Noam Shazeer, Niki Parmar, Jakob Uszkoreit, Llion Jones,
  Aidan~N Gomez, {\L}ukasz Kaiser, and Illia Polosukhin. 2017.
\newblock Attention {I}s {A}ll {Y}ou {N}eed.
\newblock In \emph{Advances in neural information processing systems}, pages
  5998--6008.

\bibitem[{Wang et~al.(2021)Wang, Cui, and Yu}]{wang-etal-2021-bragging}
Jun Wang, Kelly Cui, and Bei Yu. 2021.
\newblock \href {https://doi.org/10.18653/v1/2021.naacl-main.388} {Self
  {P}romotion in {US} {C}ongressional {T}weets}.
\newblock In \emph{Proceedings of the 2021 Conference of the North American
  Chapter of the Association for Computational Linguistics: Human Language
  Technologies}, pages 4893--4899, Online. Association for Computational
  Linguistics.

\bibitem[{Waseem and Hovy(2016)}]{waseem-hovy-2016-hateful}
Zeerak Waseem and Dirk Hovy. 2016.
\newblock \href {https://doi.org/10.18653/v1/N16-2013} {Hateful symbols or
  hateful people? predictive features for hate speech detection on {T}witter}.
\newblock In \emph{Proceedings of the {NAACL} Student Research Workshop}, pages
  88--93, San Diego, California. Association for Computational Linguistics.

\bibitem[{Wen et~al.(2017)Wen, Wang, Dong, and Chen}]{wen2017jointly}
Liyun Wen, Xiaojie Wang, Zhenjiang Dong, and Hong Chen. 2017.
\newblock Jointly {M}odeling {I}ntent {I}dentification and {S}lot {F}illing
  with {C}ontextual and {H}ierarchical {I}nformation.
\newblock In \emph{National CCF Conference on Natural Language Processing and
  Chinese Computing}, pages 3--15. Springer.

\bibitem[{Wittels(2011)}]{wittels2011humblebrag}
Harris Wittels. 2011.
\newblock Humblebrag hall of fame.
\newblock \emph{Grantland. com}.

\bibitem[{Yang et~al.(2021)Yang, Hooshmand, and
  Hirschberg}]{yang-etal-2021-choral}
Zixiaofan Yang, Shayan Hooshmand, and Julia Hirschberg. 2021.
\newblock \href {https://aclanthology.org/2021.emnlp-main.364} {{CH}o{R}a{L}:
  Collecting humor reaction labels from millions of social media users}.
\newblock In \emph{Proceedings of the 2021 Conference on Empirical Methods in
  Natural Language Processing}, pages 4429--4435, Online and Punta Cana,
  Dominican Republic. Association for Computational Linguistics.

\end{thebibliography}
\bibliographystyle{acl_natbib}

\clearpage 

\appendix


\section{Impact of Multiple Annotations}
\label{s:annotation_results}

Table \ref{t:subset_binary} shows the performance of binary bragging classification of the best performing model (BERTweet-LIWC) on two different subsets of the test data: one annotated by a single annotator (2,130 tweets) and the other annotated by two or more annotators until consensus is reached (522 tweets). The results show that the same model tested on the two different subsets of test data lead to similar results. This shows there is no quantitative difference between the data sets annotated by two or more annotators when compared to a single annotator.

\renewcommand{\arraystretch}{1.1}
\begin{table}[!h]
\small
\center
\begin{tabular}{|l|c|c|c|}
\hline
\rowcolor{darkGray} \bf Data set & \bf Precision & \bf Recall & \bf Macro-F1\\
\hline
Single Annotation & 73.81 & 71.78 & 72.74 \\
\hline
Multiple Annotations & 68.24 & 83.31 & 73.23 \\
\hline
Entire set & 72.92 & 72.81 & 72.86 \\
\hline
\end{tabular}
\caption{Precision, Recall and macro F1-Score obtained by the same best performing model (BERTweet-LIWC) for binary classification on two different subsets of training data, annotated either by a single annotator or by multiple annotators.}
\label{t:subset_binary}
\end{table}


\begin{figure}[!h]
\center
\includegraphics[scale=0.45]{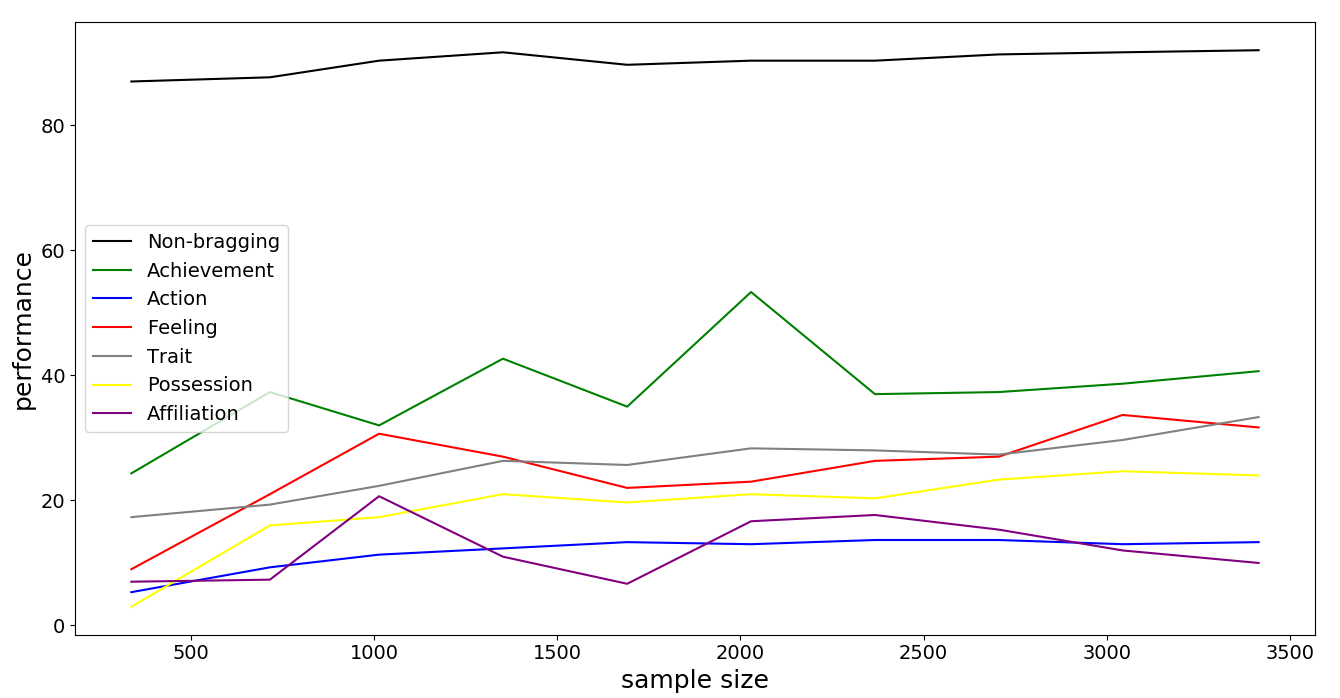}
\caption{Learning curve for performance across each bragging type.}             
\label{fig:learning_curve} 
\end{figure}

\section{Guidelines and Annotation Interface}
\label{s:interface}

\begin{figure}[!t]
\center
\includegraphics[scale=0.95]{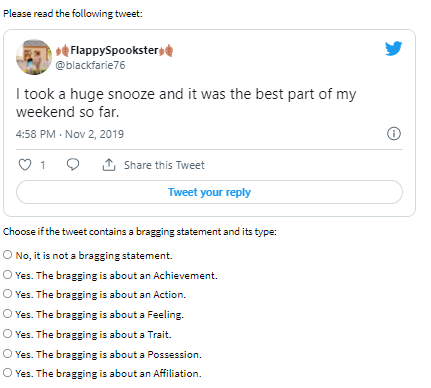}
\caption{Screenshot of annotation interface on our platform.}    
\label{fig:interface} 
\end{figure}

Thank you for your participation in our study. During our experiment, we will ask you to read and evaluate a tweet which may include a bragging or a praisal statement.

\paragraph{Instructions}

You need to identify whether or not a tweet includes a bragging statement.

\paragraph{Bragging}~Bragging is a speech act which explicitly or implicitly attributes credit to the speaker for some ‘good’ (possession, accomplishment, skill, etc.) which is positively valued by the speaker and the potential audience.
As such, bragging includes announcements of accomplishments, explicit positive evaluations of some aspect of self and other types defined below.
A bragging statement should clearly express what the author is bragging about (i.e. the target of bragging).

If the tweet is about bragging, decide on the category where the tweet belongs to from the following categories:

\paragraph{Achievement}~The act of bragging is about a concrete outcome obtained as a result of the tweet author’s actions. These results may include achievements, awards, products, and/or positive change in a situation or status (individually or as part of a group).

Examples:
\begin{itemize}[noitemsep,topsep=0pt,leftmargin=1em]
\item \textit{Finally got that offer! Whoop!!}
\item \textit{Our team won the championship}
\end{itemize}

\paragraph{Action}~The act of bragging is about a past, current or upcoming action of the user that does not have a concrete outcome

Examples:
\begin{itemize}[noitemsep,topsep=0pt,leftmargin=1em]
\item \textit{Hanging at Buffalo Wild Wings with @user for the \#ILLvsASU game. \#BraggingRights}
\item \textit{Guess what! I met Matt Damon today!}
\end{itemize}

\paragraph{Feeling}~The act of bragging is about a feeling that is expressed by the user for a particular situation.

Example:
\begin{itemize}[noitemsep,topsep=0pt,leftmargin=1em]
\item \textit{Im so excited that I am back on my consistent schedule. I am so excited for a routine so I can achieve my goals!!}
\end{itemize}

\paragraph{Trait}~The act of bragging is about a personal trait, skill or ability of the user . 

Examples:
\begin{itemize}[noitemsep,topsep=0pt,leftmargin=1em]
\item \textit{To be honest, I have a better memory than my siblings}
\item \textit{I look great after losing weight}
\end{itemize}

\paragraph{Possession}~The act of bragging is about a tangible object belonging to  the user. 

Example:
\begin{itemize}[noitemsep,topsep=0pt,leftmargin=1em]
\item \textit{Look at our Christmas tree! I kinda just wanna keep it up all year!}
\end{itemize}

\paragraph{Affiliation}~The act of bragging is about being part of a group  (e.g. family, team, org etc.) and/or a certain location including living in a city, neighborhood or country, enrolled into a university, supporting a team, working in a company etc.

Example:
\begin{itemize}[noitemsep,topsep=0pt,leftmargin=1em]
\item \textit{My daughter got first place in the final exam, so proud of her!}
\end{itemize}

\paragraph{Not bragging}~If the tweet is not about bragging, then select "No. This is not a bragging statement."

Examples:
\begin{itemize}[noitemsep,topsep=0pt,leftmargin=1em]
\item \textit{One of the best books I've ever read}
\item \textit{hahahahahaha}
\item \textit{You gotta admit, that's some mighty awesome aim!}
\item \textit{Vote in the poll below for your book of choice!}
\item \textit{I think this is great}
\item \textit{dear everyone announcing they are at "Friendsgiving", we get it, you have friends}
\item \textit{In case you didn't know, Adam Silver is in charge}
\item \textit{I feel terrible}
\item \textit{I don't know why you are celebrating}
\item \textit{This is exactly what is going on!}
\item \textit{I love you}
\end{itemize}

Select "No. This is not a bragging statement", also in cases when:
\begin{itemize}[noitemsep,topsep=0pt,leftmargin=1em]
\item there is not enough information to determine that the tweet is about bragging
\item the bragging statements belong to someone other than the author of the tweet
\item the relationship between author and people/things mentioned in the tweet are unknown:
\begin{itemize}[noitemsep,topsep=0pt,leftmargin=1em]
\item \textit{This kid is smart}
\item \textit{That was an amazing stream}
\item \textit{Kudos to mike Dunleavy! It's hard to get a franchise record ANYTHING in Chicago}
\end{itemize}
\item the post is about the act of bragging:
\begin{itemize}[noitemsep,topsep=0pt,leftmargin=1em]
\item \textit{We want to hear you brag!}
\item \textit{Trump isn't Bragging anymore as his tradewar hits the stockmarket hard}
\item \textit{Dudes are getting too cocky these days. Them lil labels and that dar don't impress everyone. brag differently}
\end{itemize}
\end{itemize}

\paragraph{Not available}~Finally, if the tweet is not available or displayed, or is in a language other than English, please select the "Not available" option.

\paragraph{Other considerations}

Please verify the content of hashtags as these may give clues towards the category of the tweet. The judgment should be made only based on the given content of the tweet - please do not search the tweet on Twitter or online in order to identify additional context.


\begin{figure}[!t]
\center
\includegraphics[scale=0.4]{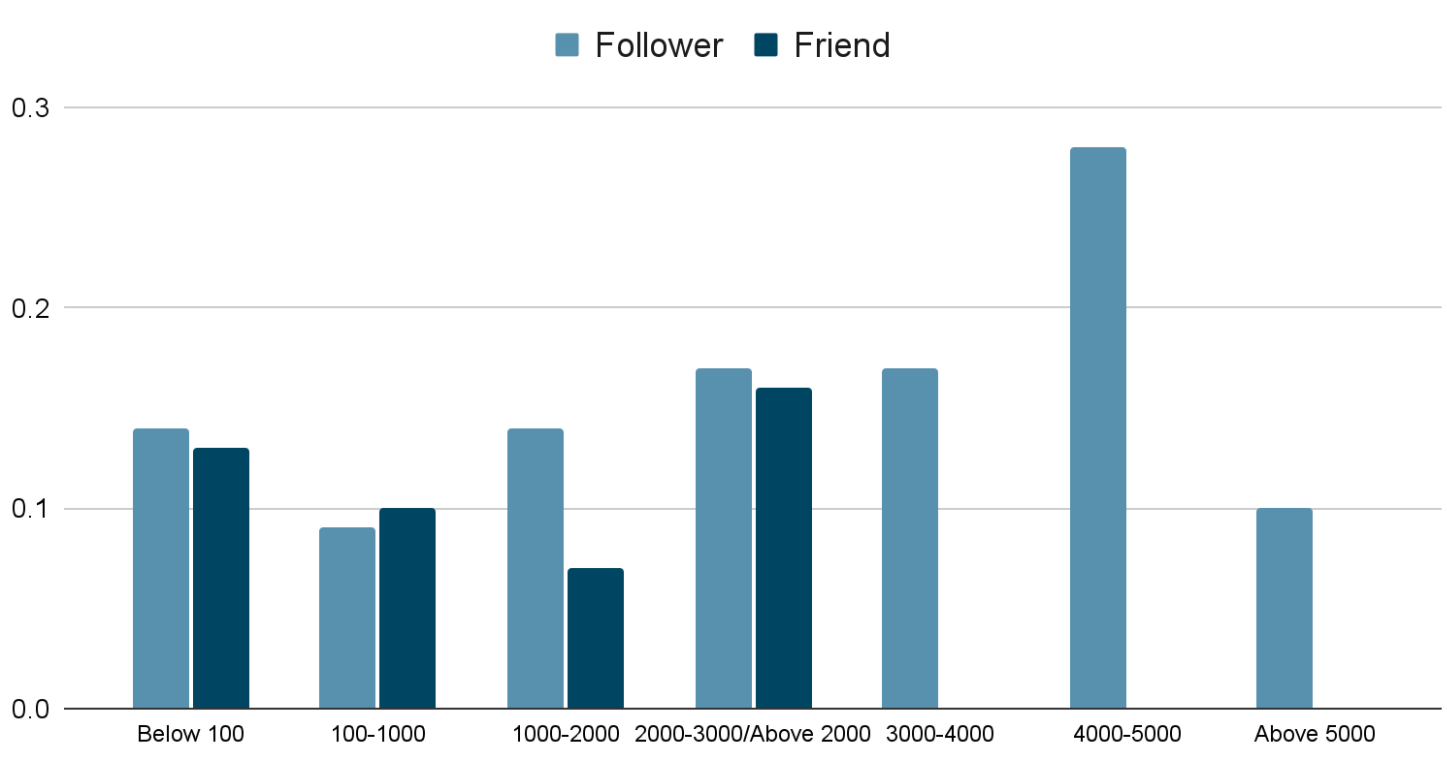}
\caption{Pearson correlation between Twitter favorite number and bragging by controlling the number of followers and friends. All correlations are significant at $p$ < .01, two-tailed t-test.} \label{fig:correlation} 
\end{figure}


\end{document}